\pdfoutput=1

\documentclass[11pt]{article}

\usepackage{EMNLP2022}

\usepackage{times}
\usepackage{latexsym}

\usepackage[T1]{fontenc}

\usepackage[utf8]{inputenc}

\usepackage{microtype}

\usepackage{inconsolata}

\usepackage{xspace}
\usepackage{booktabs}
\usepackage{caption}
\usepackage{subcaption}
\usepackage{graphicx}
\usepackage{float}
\usepackage[normalem]{ulem}
\usepackage{multirow}
\usepackage{amsmath}
\usepackage{enumitem}

\usepackage{balance}

\usepackage{mkolar_definitions}

\newcommand{\cls}{\textsc{cls}\xspace}
\newcommand{\nystrom}{Nyström\xspace}
\newcommand{\tfidf}{\textsc{tf-idf}\xspace}
\newcommand{\zeshel}{\textsc{ZeShEL}\xspace}

\newcommand{\clsCrossenc}{\textsc{[cls]-CE}\xspace}
\newcommand{\eCrossenc}{\textsc{[emb]-CE}\xspace}

\newcommand{\yugioh}{\texttt{YuGiOh}\xspace}
\newcommand{\starTrek}{\texttt{Star\_Trek}\xspace}
\newcommand{\doctorWho}{\texttt{Doctor\_Who}\xspace}
\newcommand{\proWrestling}{\texttt{Pro\_Wrestling}\xspace}
\newcommand{\proWrestlingShort}{\texttt{Pro\_Wrest}\xspace}
\newcommand{\military}{\texttt{Military}\xspace}

\newcommand{\query}{q\xspace}
\newcommand{\querySpace}{\mathcal{Q}\xspace}
\newcommand{\testQuery}{q_{\texttt{test}}\xspace}
\newcommand{\dataItem}{i\xspace}
\newcommand{\itemSpace}{\mathcal{I}\xspace}
\newcommand{\anchorItems}{\mathcal{I}_{\texttt{anc}}\xspace}
\newcommand{\anchorQueries}{\mathcal{Q}_{\texttt{anc}}\xspace}

\newcommand{\rowMatrixTrain}{R_\text{anc}\xspace}

\newcommand{\colMatrixTrain}{C_\text{anc}\xspace}
\newcommand{\colMatrixTest}{C_\text{test}\xspace}
\newcommand{\augmentedMatrixCols}{C_\text{comb}\xspace}

\newcommand{\augmentedMatrix}{M_\text{comb}\xspace}
\newcommand{\approxAugmentedMatrix}{\Tilde{M}_\text{comb}\xspace}

\newcommand{\matrixTrain}{M_\text{anc}\xspace}
\newcommand{\matrixTest}{M_\text{test}\xspace}
\newcommand{\approxMatrix}{\Tilde{M}\xspace}
\newcommand{\approxMatrixTrain}{\approxMatrix_\text{anc}\xspace}
\newcommand{\approxMatrixTest}{\approxMatrix_\text{test}\xspace}
\newcommand{\pseudoInvColMatrixTrain}{C^\dagger_\text{anc}}

\newcommand{\itemRep}{E^{\itemSpace}\xspace}

\newcommand{\nAnchorQueries}{k_q\xspace}
\newcommand{\nAnchorItems}{k_i\xspace}
\newcommand{\nAnchorItemsForIndex}{k_i^{\text{ind}}\xspace}
\newcommand{\nItems}{\lvert \itemSpace \rvert\xspace}
\newcommand{\nRetrievedItems}{k_r\xspace}

\newcommand{\queryEmbed}{e_q\xspace}
\newcommand{\cost}{\mathcal{C}}

\newcommand{\model}[1]{f_{#1}\xspace}

\newcommand{\queryEmbedFromCE}{e^{\textsc{ce}}_q}
\newcommand{\itemEmbedFromCE}{e^{\textsc{ce}}_i}
 
\newcommand{\bert}{\textsc{Bert}\xspace}
\newcommand{\trans}{\mathcal{T}}

\newcommand{\fixedDualEncoder}{\textsc{DE\textsubscript{base}}\xspace}
\newcommand{\finetuneDualEncoder}{\textsc{DE\textsubscript{base+ce}}\xspace}
\newcommand{\scratchDualEncoder}{\textsc{DE\textsubscript{bert+ce}}\xspace}

\newcommand{\propMethod}{\textsc{annCUR}\xspace}
\newcommand{\fixedItems}{\textsc{fixedITEM}\xspace}
\newcommand{\itemCUR}{\textsc{itemCUR}\xspace}

\newcommand{\queryTestData}{\mathcal{Q}_\text{test}}
\newcommand{\queryTrainData}{\mathcal{Q}_\text{train}}
\newcommand{\queryTrainSize}{\lvert \queryTrainData \rvert}
\newcommand{\queryItemCompTime}{t_\text{CE-Mat}}
\newcommand{\dualEncTrainTime}{t_\text{DE-train}}

\newcommand{\kDistill}{k_d}
\newcommand{\topKEnts}[3]{\textsc{T}^{\text{#2}}_{#1}(#3)\xspace}
\newcommand{\topKNegEnts}[3]{\textsc{N}^{\text{#2}}_{#1}(#3)\xspace}

\newcommand{\ment}{q\xspace}

\newcommand{\loss}[1]{\mathcal{L}_\text{#1}}
\newcommand{\crossEntropy}{\mathcal{H}}
\newcommand{\softmax}{\sigma}
\newcommand{\scoreMat}[1]{S^{\text{(#1)}}}

\newcommand{\posNegPairs}[1]{\mathcal{P}_{#1}}

\newcount\Comments  
\definecolor{purple}{rgb}{0.5,0,1}
\definecolor{teal}{rgb}{0.33,0.65,0.55}
\definecolor{green}{rgb}{0.1,0.65,0.1}
\newcommand{\kibitz}[2]{\ifnum\Comments=1\textcolor{#1}{#2}\fi}

\title{Efficient Nearest Neighbor Search for Cross-Encoder Models \\
using Matrix Factorization}

\author{Nishant Yadav$^\dag$, Nicholas Monath$^{\diamondsuit}$,  Rico Angell$^\dag$, \\  \textbf{Manzil Zaheer}$^{\diamondsuit}$,  and \textbf{Andrew McCallum}$^\dag$ \\
$^\dag$ University of Massachusetts Amherst \quad $^{\diamondsuit}$Google Research \\
\texttt{\{nishantyadav, rangell, mccallum\}@cs.umass.edu}\\ \quad \texttt{\{nmonath,manzilzaheer\}@google.com}
}

\begin{document}
\maketitle

\begin{abstract}
    
Efficient $k$-nearest neighbor search is a fundamental task, foundational for many problems in NLP.  When the similarity is measured by dot-product between dual-encoder vectors or $\ell_2$-distance, there already exist many scalable and efficient search methods.  But not so when similarity is measured by more accurate and expensive black-box neural similarity models, such as cross-encoders, which jointly encode the query and candidate neighbor.  
The cross-encoders' high computational cost typically limits their use to reranking candidates retrieved by a cheaper model, such as dual encoder or \tfidf.  
However, the accuracy of such a two-stage approach is upper-bounded 
by the recall of the initial candidate set, and potentially requires 
additional training to align the auxiliary retrieval model with 
the cross-encoder model. 
In this paper, we present an approach that avoids the use of a dual-encoder for retrieval, relying solely on the cross-encoder.  Retrieval is made efficient with CUR decomposition, a matrix decomposition approach that approximates all pairwise cross-encoder distances from a small subset of rows and columns of the distance matrix.
Indexing items using our approach is computationally cheaper than training an 
auxiliary dual-encoder model through distillation. 
Empirically, for $k > 10$, our approach provides test-time 
recall-vs-computational cost trade-offs 
superior to the current widely-used methods that re-rank items 
retrieved using a dual-encoder or \tfidf.
    
\end{abstract}

\section{Introduction}
\label{sec:introduction}

\begin{figure*}[!ht]
     \centering
     \begin{subfigure}[b]{0.73\textwidth}
         \centering
         \includegraphics[width=\textwidth]{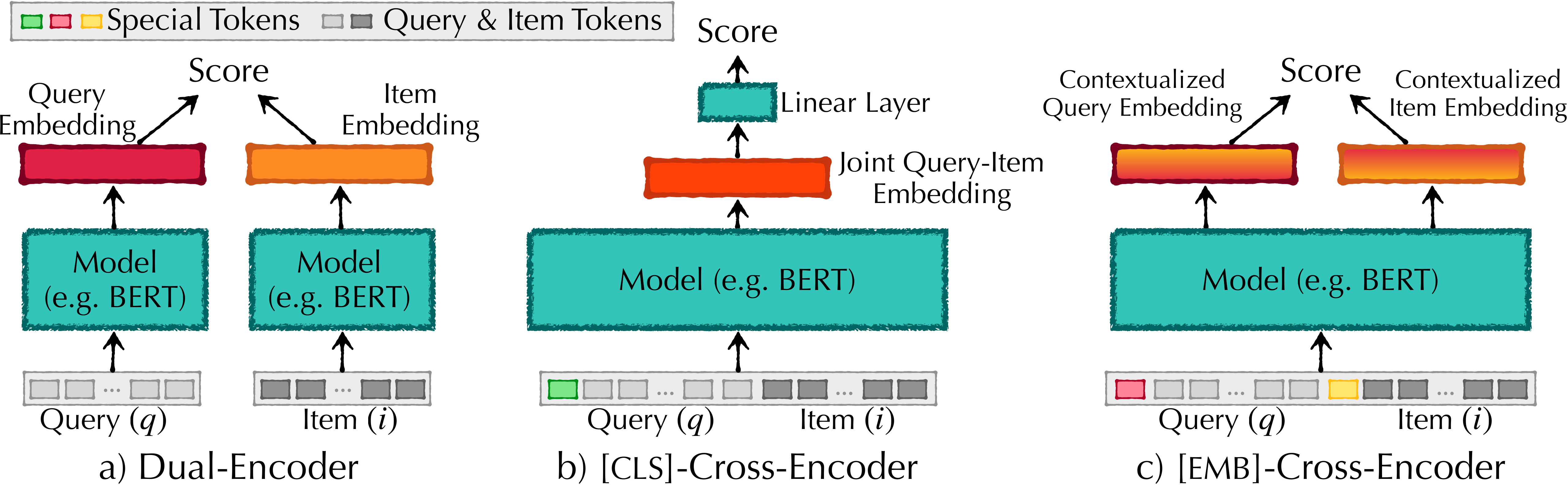}
        \caption{Model architecture}
        \label{fig:arch_diagram}
     \end{subfigure}
     \hfill
     \begin{subfigure}[b]{0.259\textwidth}
         \centering
        \includegraphics[width=\textwidth]{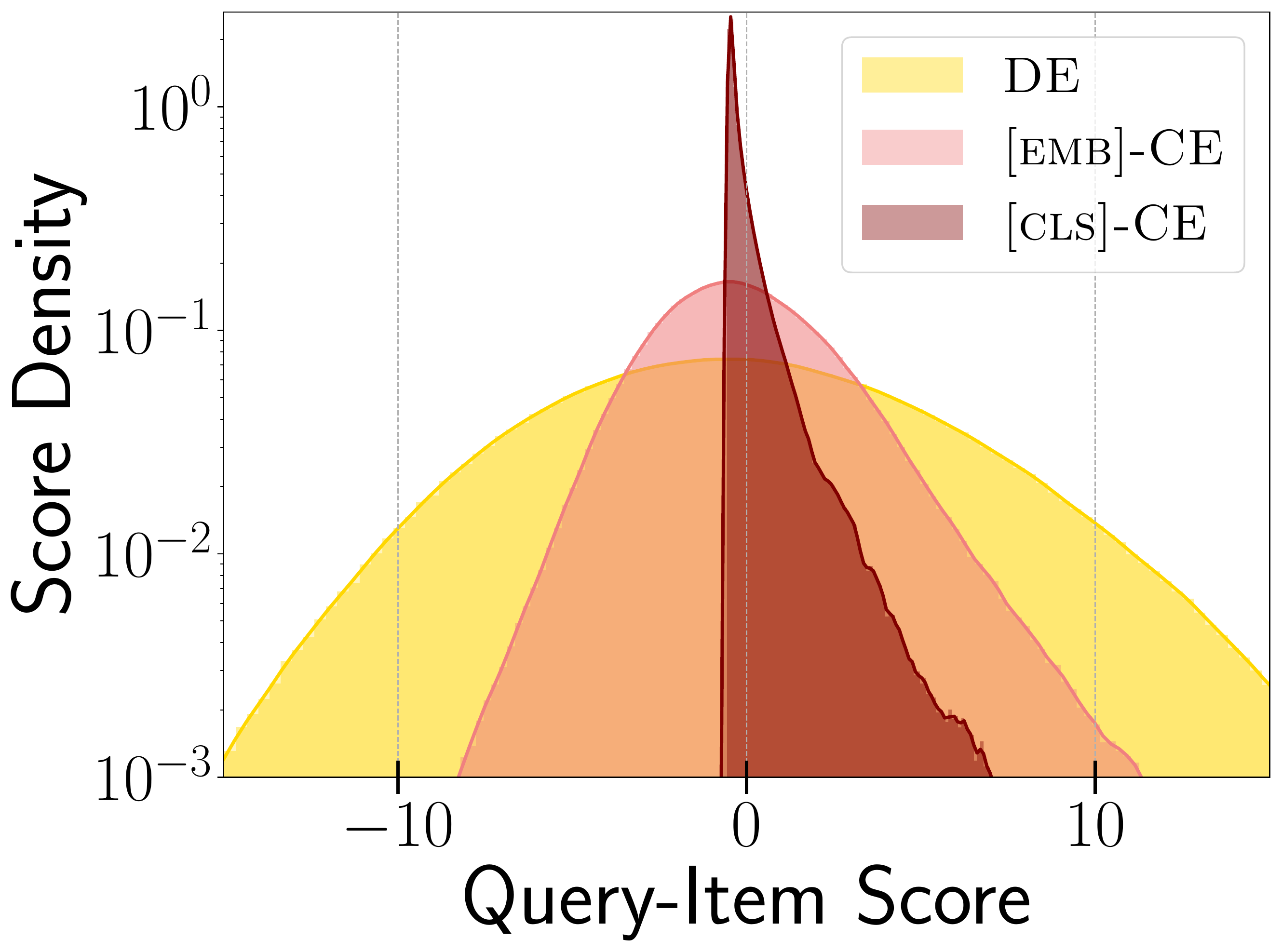}
    \caption{Query-item score distribution}
    \label{fig:score_dist_all_models}
     \end{subfigure}
     \caption{Model architecture and score distribution for three neural scoring functions. Dual-Encoder (DE) models score a query-item pair using \emph{independently} computed query and item embeddings. \clsCrossenc model computes the score by jointly encoding the query-item pair followed by passing the joint query-item embedding through a linear layer. Our proposed \eCrossenc model embeds special tokens amongst query and item tokens, and computes the query-item score using contextualixed query and item embeddings extracted using the special tokens after jointly encoding the query-item pair.}
     \vspace{-0.5cm}
\end{figure*}

Finding top-$k$ scoring items for a given query is a 
fundamental sub-routine of recommendation and
information retrieval systems~\cite{kowalski2007information, das2017survey}. 
For instance, in question answering systems, the query corresponds to 
a question and the item corresponds to a document or a passage.
Neural networks are widely used to model the 
similarity between a query and an item in such applications~\cite{zamani2018neural, hofstatter2019effect, karpukhin2020dense, qu-etal-2021-rocketqa}.
In this work, we focus on efficient $k$-nearest neighbor
search for one such similarity function -- the cross-encoder model.

Cross-encoder models output a scalar 
similarity score by jointly encoding the query-item pair and
often generalize better to new domains and unseen data~\cite{chen-etal-2020-dipair, wu-etal-2020-scalable, thakur-etal-2021-augmented}
as compared to dual-encoder~\footnote{also referred to as two-tower models, Siamese networks} 
models which independently embed the query and the item in a vector space,
and use simple functions such as dot-product to measure similarity.
However, due to the black-box nature of the cross-encoder based similarity
function, the computational cost for brute force search with cross-encoders is prohibitively high. 
This often limits the use of cross-encoder models to re-ranking items retrieved 
using a separate retrieval model such as a dual-encoder or a \tfidf-based 
model~\cite{logeswaran-etal-2019-zero,  zhang-stratos-2021-understanding, qu-etal-2021-rocketqa}.
The accuracy of such a two-stage approach is upper bounded by the recall 
of relevant items by the initial retrieval model. 
Much of recent work either attempts to distill information from an
expensive but more expressive cross-encoder
model into a cheaper student model such as a
dual-encoder~\cite{wu-etal-2020-scalable, Hofsttter2020ImprovingEN,  lu2020twinbert, qu-etal-2021-rocketqa, liu2021trans},
or focuses on cheaper alternatives to the cross-encoder model 
while attempting to capture fine-grained interactions between the query and the item
~\cite{humeau2019poly, khattab2020colbert, luan2021sparse}. 

In this work, we tackle the fundamental task of efficient $k$-nearest
neighbor search for a given query according to the cross-encoder.
Our proposed approach, \propMethod, uses CUR decomposition~\cite{mahoney2009cur}, a matrix factorization approach,
to approximate cross-encoder scores for all items,
and retrieves $k$-nearest neighbor items while only making a small
number of calls to the cross-encoder. 
Our proposed method selects a fixed set of anchor queries 
and anchor items, and uses scores between anchor queries and all items to 
generate latent embeddings for indexing the item set. 
At test time, we generate latent embedding for the query
using cross-encoder scores for the test query and anchor items,
and use it to approximate scores 
of all items for the given query and/or retrieve top-$k$ 
items according to the approximate scores. 
In contrast to distillation-based approaches, our proposed
approach does not involve any additional compute-intensive
training of a student model such as dual-encoder via distillation. 

In general, the performance of a matrix factorization-based method 
depends on the rank of the matrix being factorized.
In our case, the entries of the matrix are cross-encoder scores for query-item
pairs. To further improve rank of the score matrix, and in-turn performance of the proposed matrix factorization based approach,
we propose \eCrossenc which uses a novel dot-product based scoring mechanism for cross-encoder models (see Figure~\ref{fig:arch_diagram}).
In contrast to the widely used \clsCrossenc approach of pooling query-item
representation into a single vector followed by scoring using a linear layer,
\eCrossenc produces a score matrix with a much lower rank while performing
at par with \clsCrossenc on the downstream task. 

We run extensive experiments with cross-encoder models
trained for the downstream task of entity linking. The query
and item in this case correspond to a mention of an entity in text
and a document with an entity description respectively.
For the task of retrieving $k$-nearest neighbors according to the cross-encoder,
our proposed approach presents superior recall-vs-computational cost 
trade-offs over using dual-encoders trained via 
distillation as well as over unsupervised \tfidf-based methods (\S\ref{subsec:exp_cur_vs_baselines}).
We also evaluate the proposed method for various indexing and 
test-time cost budgets as well as study the effect of
various design choices in~\S\ref{subsec:cur_analysis} and~\S\ref{subsec:item_ce_baselines}.

\section{Matrix Factorization for Nearest Neighbor Search}
\label{sec:model}

\subsection{Task Description and Background}
\label{subsec:task_description}

\begin{figure*}[!ht]
    \centering
    \includegraphics[width=0.95\textwidth]{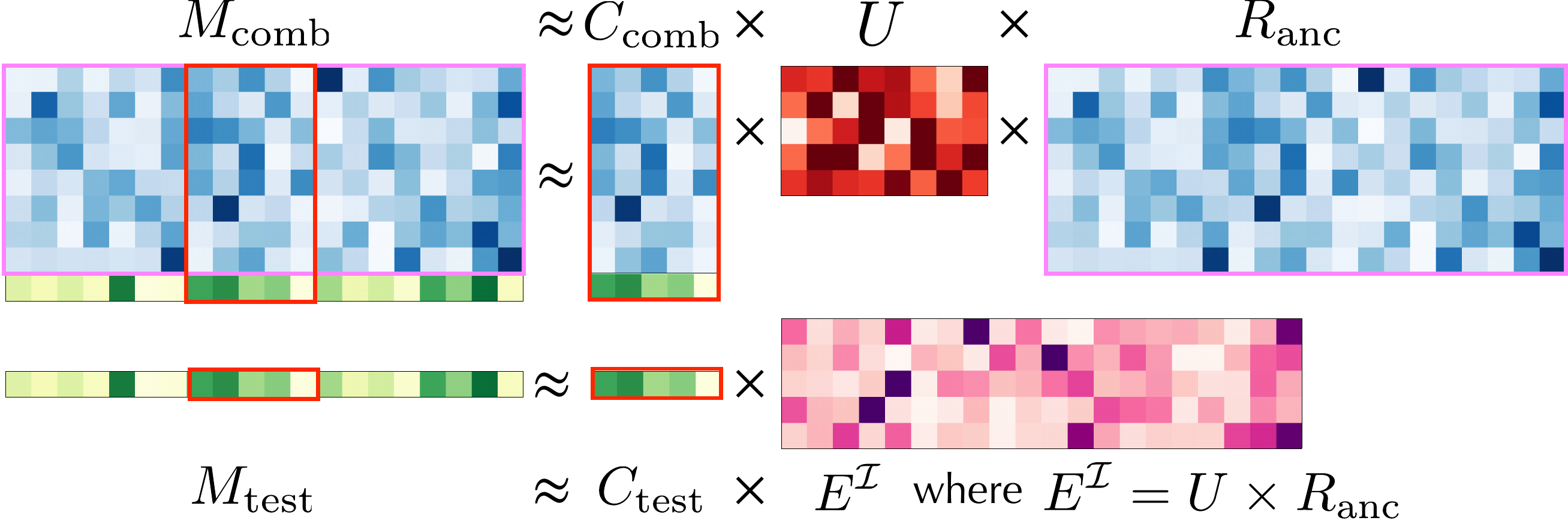}
    \caption{CUR decomposition of a matrix 
    ($\augmentedMatrix = \begin{bmatrix} \matrixTrain \\  \matrixTest \\ \end{bmatrix}$) 
    using a subset of its columns 
    ($\augmentedMatrixCols = \begin{bmatrix} \colMatrixTrain \\ \colMatrixTest \end{bmatrix}$) 
    and rows ($\rowMatrixTrain$). The blue rows ($\rowMatrixTrain$) 
    corresponding to the anchor queries and $U = \colMatrixTrain^\dagger$
    are used for indexing the items to obtain item embeddings
    $\itemRep = U \times \rowMatrixTrain$ and the green row 
    corresponds to the test query. Note that the test row ($\matrixTest$) can be approximated 
    using a subset of its columns ($\colMatrixTest$) and the latent
    representation of items ($\itemRep$).} 
    \label{fig:matrix_factorization}
    \vspace{-0.5cm}
\end{figure*}

Given a scoring function
$\model{\theta} : \querySpace \times \itemSpace \rightarrow \RR $ 
that maps a query-item pair to a scalar score, and a
query $\query \in \querySpace$, the $k$-nearest neighbor task is to retrieve top-$k$
scoring items according to 
the given scoring function $\model{\theta}$
from a fixed item set $\itemSpace$.

In NLP, queries and items are typically represented as a sequence of 
tokens and the scoring function is typically parameterized using 
deep neural models such as transformers~\cite{vaswani2017attention}.
There are  two popular choices for the scoring function -- 
the cross-encoder (CE) model, and the  dual-encoder (DE) model.
The CE model  
scores a given query-item pair by concatenating the query and the item 
using special tokens, passing them through a model (such as a transformer $\trans$)
to obtain representation for the input pair followed by 
computing the score using linear weights $w \in \RR^{d}$.
\vspace{-0.25cm}
\begin{equation*}
\model{\theta}^{(\text{CE})}(\query, \dataItem) = w^{\top} \texttt{pool}(\trans(\texttt{concat}(\query, \dataItem)))
\vspace{-0.25cm}
\end{equation*}
While effective, computing a similarity between a query-item
pair requires a full forward pass of the model, which is often
quite computationally burdensome. 
As a result, previous work uses auxiliary retrieval models such as BM25~\cite{robertson1995okapi} or a trained
 dual-encoder (DE) model to approximate the CE.
The DE model \emph{independently} embeds the query and the item in $\RR^{d}$,
for instance by using a transformer $\trans$ followed by pooling the final layer 
representations into a single vector (e.g. using \cls token).
The DE score for the query-item pair is computed using dot-product of the query embedding and the item embedding.
\[ \model{\theta}^{\text{(DE)}}(\query, \dataItem) = \texttt{pool}(\trans(\query))^{\top} \texttt{pool}(\trans(\dataItem)) \]
In this work, we propose a method based on CUR
matrix factorization that allows efficient retrieval
of top-$k$ items by directly approximating the cross-encoder model rather than using an auxiliary (trained) retrieval model.

\paragraph{CUR Decomposition}~\cite{mahoney2009cur} 
In CUR matrix factorization,  a matrix $M \in \RR^{n \times m}$ is approximated using a subset of its rows $R = M[S_r, :] \in \RR^{k_1 \times m}$,
a subset of its columns $C = M[:, S_c] \in \RR^{n \times k_2}$ and a joining
matrix $U \in \RR^{k_2 \times k_1}$ as follows
\vspace{-0.25cm}
\begin{equation*}
    \Tilde{M} = C U R 
    \vspace{-0.25cm}
\end{equation*} 

where $S_r$ and $S_c$ are the indices corresponding to rows $R$ and columns $C$ respectively,
and the joining matrix $U$ optimizes the approximation error. 
In this work, we set $U$ to be the Moore-Penrose pseudo-inverse
of $M[S_r, S_c]$, the intersection of  matrices $C$ and $R$,
in which case $\Tilde{M}$ is known as
the skeleton approximation of $M$~\cite{goreinov1997theory}.

\subsection{Proposed Method Overview}
\label{subsec:proposed_method}

Our proposed method \propMethod, which stands
for \textbf{A}pproximate \textbf{N}earest \textbf{N}eighbor search using \textbf{CUR} decomposition,
begins by selecting a fixed set of $\nAnchorQueries$ \emph{anchor} queries
($\anchorQueries$) and $\nAnchorItems$ \emph{anchor} items ($\anchorItems$).
It uses 
scores between queries $\query \in \anchorQueries$
and all items $\dataItem \in \itemSpace$ to index the item set
by generating latent item embeddings. 
At test time, we compute  exact scores between the test query $\testQuery$ and the anchor items $\anchorItems$,
and use it to approximate scores of all items for the given query and/or
retrieve top-$k$ items according to the approximate scores.
We could optionally retrieve $\nRetrievedItems > k$ items, re-rank them
using exact $\model{\theta}$ scores and return top-$k$ items.

Let $\augmentedMatrix = \begin{bmatrix} \matrixTrain \\  \matrixTest \\ \end{bmatrix}$ and $\augmentedMatrixCols = \begin{bmatrix} \colMatrixTrain \\ \colMatrixTest \end{bmatrix} $
where $ \matrixTrain = \rowMatrixTrain \in \RR^{\nAnchorQueries \times \nItems}$ contains scores for 
the anchor queries and all items,
$ \matrixTest \in \RR^{1 \times \nItems}$ contains 
scores for a test query and all items,
$\colMatrixTrain \in \RR^{\nAnchorQueries \times \nAnchorItems}$ contains scores for the anchor
queries and the anchor items, and $\colMatrixTest \in \RR^{1 \times \nAnchorItems}$ contains 
scores for the test query paired with the anchor items.

Using CUR decomposition, we can approximate $\augmentedMatrix$ using a subset of its columns ($\augmentedMatrixCols$) corresponding
to the anchor items and a subset of its
rows ($\rowMatrixTrain$) corresponding to the anchor queries as
\vspace{-0.6cm}
\begin{align*}
   \approxAugmentedMatrix &= \augmentedMatrixCols U \rowMatrixTrain  \\
   \begin{bmatrix} \approxMatrixTrain \\  \approxMatrixTest \\ \end{bmatrix} &= \begin{bmatrix} \colMatrixTrain \\ \colMatrixTest \end{bmatrix} U \rowMatrixTrain \\
   \approxMatrixTrain =  \colMatrixTrain U \rowMatrixTrain \text{  } & \text{ and }  \text{  }  \approxMatrixTest =  \colMatrixTest U \rowMatrixTrain 
   \vspace{-0.4cm}
\end{align*}
Figure~\ref{fig:matrix_factorization} shows CUR decomposition of matrix $\augmentedMatrix$.
At test time, $\approxMatrixTest$ containing approximate item scores for the test query   
can be computed using $\colMatrixTest, U$, and $\rowMatrixTrain$ where $\colMatrixTest$ contains exact $\model{\theta}$ 
scores between the test query and the anchor items. Matrices $U$ and $\rowMatrixTrain$ can be computed offline 
as these are independent of the test query.

\subsection{Offline Indexing}
\label{subsec:indexing}
The indexing process first computes $\rowMatrixTrain$
containing scores between the anchor queries and all items.
\vspace{-0.2cm}
\begin{equation*}
    \rowMatrixTrain(q,i) = \model{\theta}(q, i), \hfill \text{ }  \forall (q,i) \in \anchorQueries \times \itemSpace
    \vspace{-0.1cm}
\end{equation*} 
We embed all items in $\RR^{\nAnchorItems}$ as
\vspace{-0.2cm}
\begin{equation*}
   \itemRep = U \times \rowMatrixTrain 
   \vspace{-0.2cm} 
\end{equation*}
where $U = \pseudoInvColMatrixTrain$ is the pseudo-inverse of $\colMatrixTrain$.
Each column of $\itemRep \in \RR^{\nAnchorItems \times \nItems}$ corresponds to a
latent item embedding.

\subsection{Test-time Inference}
\label{subsec:inference}
At test time, we embed the test query $\query$ in $\RR^{\nAnchorItems}$
using scores between $\query$ and anchor items $\anchorItems$.
\vspace{-0.2cm}
\begin{equation*}
    \queryEmbed = [ \model{\theta}(\query, \dataItem) ]_{\dataItem \in \anchorItems}
    \vspace{-0.1cm} 
\end{equation*}
We approximate the score for a query-item pair ($\query, \dataItem$) using
inner-product of $\queryEmbed$ and $\itemRep[:,\dataItem]$ where $\itemRep[:,\dataItem] \in \RR^{\nAnchorItems}$
is the embedding of item $i$.
\vspace{-0.2cm}
\begin{equation*}    
    \hat{\model{\theta}}(q,i) = \queryEmbed^{\top} \itemRep[:,\dataItem]
    \vspace{-0.1cm}
\end{equation*}

We can use $\queryEmbed \in \RR^{\nAnchorItems} $ along with 
an off-the-shelf nearest-neighbor
search method for maximum inner-product search~\cite{malkov2018efficient, johnson2019billion,guo2020accelerating}
and retrieve top-scoring items for the given query $q$ 
according to the approximate query-item scores \emph{without}
explicitly approximating scores for all the items.

\subsection{Time Complexity}
\label{subsec:time_complexity}
During indexing stage, we evaluate 
$\model{\theta}$ for $\nAnchorQueries\nItems $
query-item pairs, and compute the pseudo-inverse of 
a $\nAnchorQueries \times \nAnchorItems$
matrix. The overall time complexity of the indexing stage is 
$\mathcal{O}( \cost_{\model{\theta}} \nAnchorQueries \nItems  + \cost_\text{inv}^{\nAnchorQueries, \nAnchorItems} )$, 
where  $\cost_\text{inv}^{\nAnchorQueries, \nAnchorItems}$ is the cost of computing
the pseudo-inverse of a $k_q \times k_i$ matrix, 
and $\cost_{\model{\theta}}$ is the cost of 
computing $\model{\theta}$ on a query-item pair.
For CE models used in this work, we observe that 
$\cost_{\model{\theta}} \nAnchorQueries \nItems \gg \cost_\text{inv}^{\nAnchorQueries, \nAnchorItems}$.

At test time, we need to compute $\model{\theta}$ 
for $\nAnchorItems$ query-item pairs
followed by optionally re-ranking $\nRetrievedItems$ items 
retrieved by maximum inner-product search (MIPS).
Overall time complexity for inference is 
$\mathcal{O}\Big( (\nAnchorItems + \nRetrievedItems)\cost_{\model{\theta}} + \cost_\textsc{mips}^{\nItems, \nRetrievedItems} \Big)$,
where $\mathcal{O}(\cost_\textsc{mips}^{\nItems, \nRetrievedItems})$ is the time complexity 
of MIPS over $\nItems$ items to retrieve top-$\nRetrievedItems$ items.

\subsection{Improving score distribution of CE models for matrix factorization}
\label{subsec:ecrossenc}

The rank of the query-item score matrix, and in turn, the approximation
error of a matrix factorization method depends on the scores in the matrix.
Figure~\ref{fig:score_dist_all_models}
shows a histogram of query-item score distribution 
(adjusted to have zero mean)
for a dual-encoder and  \clsCrossenc model.
We use \clsCrossenc to refer to a cross-encoder model 
parameterized using transformers which uses \cls token
to compute a pooled representation of the input query-item pair.
Both the models are trained for zero-shot entity linking~(see \S\ref{subsec:exp_train_for_orig_zeshel} for details).
As shown in the figure, the query-item score distribution 
for the \clsCrossenc model is significantly
skewed with only a small fraction of items (entities) 
getting high scores while the score distribution for 
a dual-encoder model is less so as it is generated
explicitly using dot-product of query and item embeddings.
The skewed score distribution from \clsCrossenc leads to 
a high rank query-item score matrix, which results in a large
approximation error for matrix decomposition methods. 

We propose a small but important change to the scoring mechanism 
of the cross-encoder so that it yields a less skewed score 
distribution, thus making it much easier 
to approximate the corresponding query-item score matrix \emph{without}
adversely affecting the downstream task performance. 
Instead of using \cls token representation to score a 
given query-item pair, we add special tokens amongst the query and the item tokens
and extract \emph{contextualized} query and item
representations using the special tokens
after \emph{jointly} encoding the query-item pair using a model such as a transformer $\trans$. 
\vspace{-0.2cm}
\[ \queryEmbedFromCE, \itemEmbedFromCE =  \texttt{pool}(\trans(\texttt{concat}(\query, \dataItem)))\]
The final score for the given query-item pair is computed 
using dot-product of the contextualized query and item embeddings.
\vspace{-0.2cm}
\[ \model{\theta}^{(\eCrossenc)}(\query, \dataItem) = (\queryEmbedFromCE)^{\top} \itemEmbedFromCE \]
We refer to this model as \eCrossenc.
Figure~\ref{fig:arch_diagram} shows high-level model architecture
for dual-encoders, \clsCrossenc and \eCrossenc model.

As shown in Figure~\ref{fig:score_dist_all_models}, 
the query-item score distribution from
an \eCrossenc model resembles that from a DE model.
Empirically, we observe that rank of the query-item score matrix 
for \eCrossenc model is much lower than the rank of a 
similar matrix computed using \clsCrossenc, thus
making it much easier to approximate using matrix 
decomposition based methods.

\section{Experiments}
\label{sec:exp}
In our experiments, we use CE models trained for 
zero-shot entity linking
on \zeshel dataset (\S\ref{subsec:exp_train_for_orig_zeshel}).
We evaluate the proposed method and various baselines
on the task of finding $k$-nearest neighbors for
cross-encoder models in~\S\ref{subsec:exp_cur_vs_baselines},
and evaluate the proposed method for various indexing and 
test-time cost budgets as well as study the effect of
various design choices in~\S\ref{subsec:cur_analysis} and~\S\ref{subsec:item_ce_baselines}. 
All resources for the paper including code for all experiments and model checkpoints is available at \href{https://github.com/iesl/anncur}{https://github.com/iesl/anncur} 

\paragraph{\zeshel Dataset}
The \textbf{Ze}ro-\textbf{Sh}ot \textbf{E}ntity \textbf{L}inking (\zeshel) 
dataset was constructed by~\citet{logeswaran-etal-2019-zero} 
from Wikia.
The task of zero-shot entity linking involves linking entity mentions in text
to an entity from a list of entities with associated descriptions. 
The dataset consists of 16 different domains with eight, four, and 
four domains in training, dev, and test splits respectively. Each domain contains 
non-overlapping sets of entities, thus at test time, mentions need to be linked 
to unseen entities solely based on entity descriptions.
Table~\ref{tab:zeshel_stats} in the appendix shows dataset statistics.
In this task, queries correspond to mentions of entities along with 
the surrounding context, and items correspond to entities with their
associated descriptions.

\subsection{Training DE and CE models on \zeshel}
\label{subsec:exp_train_for_orig_zeshel}
Following the precedent set by recent papers~\cite{wu-etal-2020-scalable, zhang-stratos-2021-understanding}, 
we first train a dual-encoder model on \zeshel training 
data using hard negatives. 
We train a cross-encoder model for the task of zero-shot entity-linking
on all eight training domains using cross-entropy loss with 
ground-truth entity and negative entities mined using the dual-encoder.
We refer the reader to Appendix~\ref{subsec:appendix_train_de_and_ce_zeshel} for more details.

\paragraph{Results on downstream task of Entity Linking}
To evaluate the cross-encoder models, we retrieve 64 entities for each 
test mention using the dual-encoder model and re-rank them using a cross-encoder model. 
The top-64 entities retrieved by the DE contain the ground-truth
entity for 87.95\% mentions in test data and 92.04\% mentions in dev data. 
The proposed \eCrossenc model achieves an average accuracy of 65.49 and 66.86
on domains in test and dev set respectively, and performs at par with the 
widely used and state-of-the-art~\footnote{We observe that our 
implementation of \clsCrossenc obtains slightly different 
results as compared to 
state-of-the-art (see Table 2 in ~\citet{zhang-stratos-2021-understanding} )
likely due to minor implementation/training differences.} 
\clsCrossenc architecture which achieves an accuracy of 65.87 and 67.67 on test
and dev set respectively.
Since \eCrossenc model performs at par with \clsCrossenc on the downstream task of entity linking,
and rank of the score matrix from \eCrossenc is much lower than that from \clsCrossenc,
we use \eCrossenc in subsequent experiments.

\subsection{Evaluating on $k$-NN search for CE} 
\label{subsec:exp_cur_vs_baselines}

\begin{figure*}[!ht]
     \centering
     \begin{subfigure}[b]{\textwidth}
        \centering
        \includegraphics[width=\textwidth]{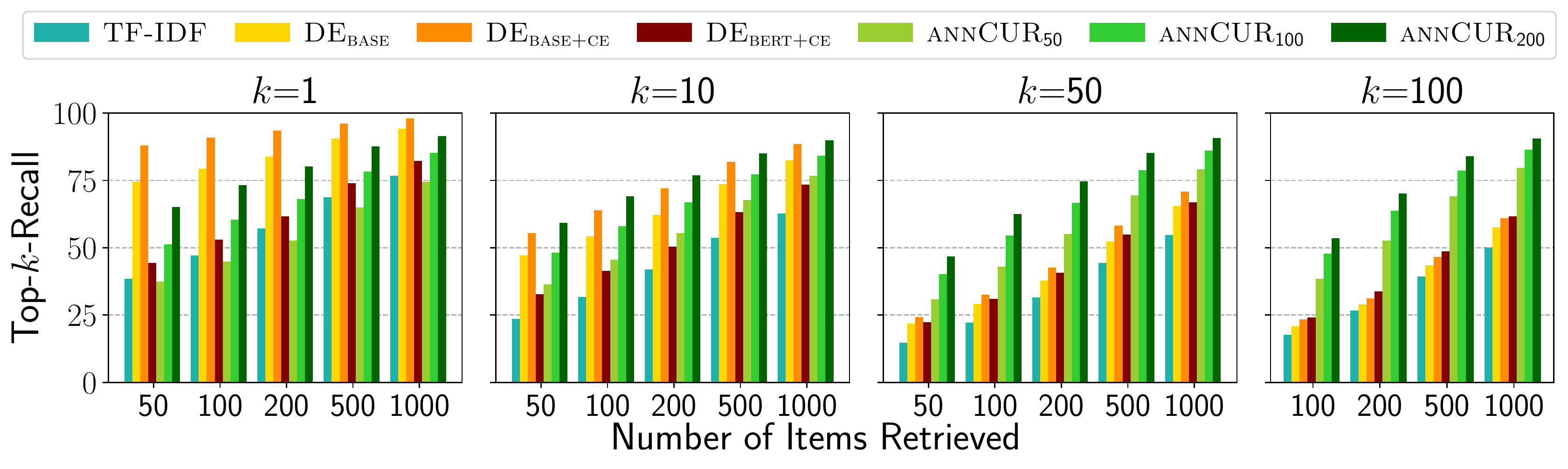}
        \caption{Top-$k$-Recall@$\nRetrievedItems$ for \propMethod and baselines when all methods retrieve and rerank 
        the same number of items $(\nRetrievedItems)$. The subscript $\nAnchorItems$ in $\propMethod_{\nAnchorItems}$ refers to
        the number of anchor items used for embedding the test query.}
        \label{fig:recall_at_same_k_retrieved_yugioh}
    \end{subfigure}
     \hfill
     \begin{subfigure}[b]{\textwidth}
    \centering
    \includegraphics[width=\textwidth]{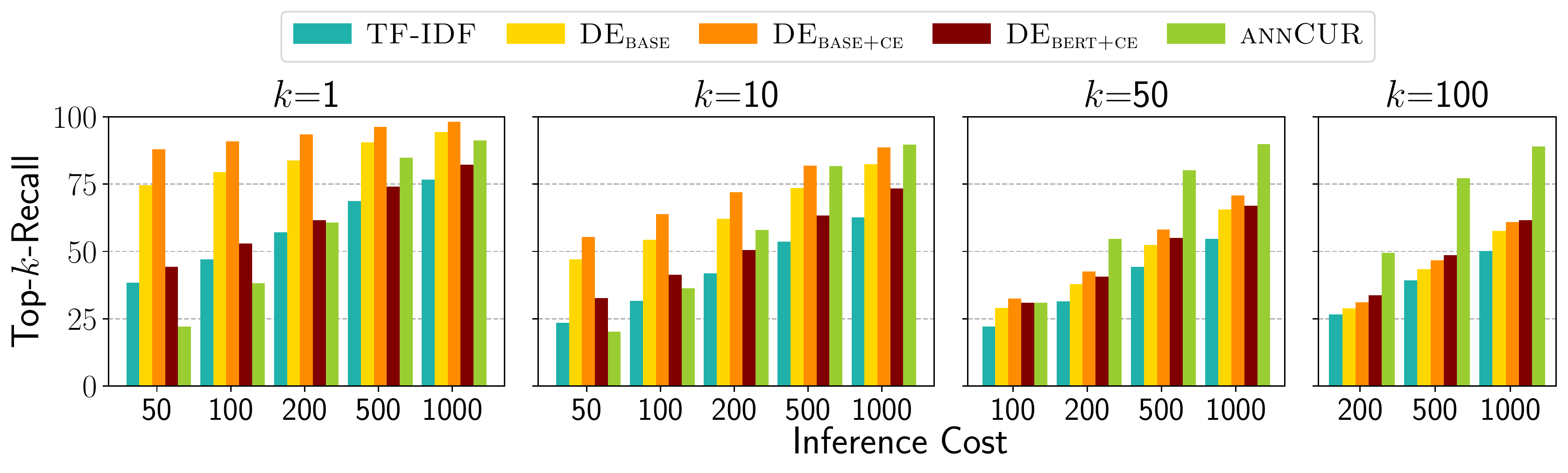}
    \caption{Top-$k$-Recall for \propMethod and baselines  when all methods operate under a fixed test-time cost budget. Recall 
    that cost is the number of CE calls made during inference
    for re-ranking retrieved items and, in case of \propMethod, it also includes CE calls to embed the test query by comparing with anchor items.
    }
    \label{fig:recall_at_same_cost_yugioh}
\end{subfigure}
    \caption{Top-$k$-Recall results for domain=\yugioh and $\queryTrainSize=500$ 
    }
    \vspace{-0.3cm}
\end{figure*}

\paragraph{Experimental Setup} 
For all experiments in this section, we use the \eCrossenc model trained
on original \zeshel training data on the task
of zero-shot entity linking, and evaluate the proposed
method and baselines for the task of retrieving $k$-nearest neighbor 
entities (items) for a given mention (query) according to the cross-encoder model. 

We run experiments \emph{separately} on five domains from \zeshel containing 10K to 100K items. 
For each domain, we compute the query-item score matrix for 
a subset or all of the queries (mentions) and all items (entities) in the domain.
We randomly split the query set into a training set ($\queryTrainData$)
and a test set ($\queryTestData$).
We use the queries in training data to train baseline DE models.
For \propMethod,  we use the training queries as anchor queries and
use CE scores between the anchor queries and all items 
for indexing as described in \S\ref{subsec:indexing}.
All approaches are then evaluated on the task of finding top-$k$ 
CE items for queries in the corresponding domain's test split.
For a fair comparison, we do not train DE
models on multiple domains at the same time.

\subsubsection{Baseline Retrieval Methods}
\label{subsec:baselines}

\paragraph{\tfidf:} All queries and items are embedded using a \tfidf vectorizer trained 
on item descriptions and top-$k$ items are retrieved using the dot-product of query and item embeddings.

\paragraph{DE models:}
We experiment with \fixedDualEncoder, the DE model 
trained on \zeshel for the task of entity linking (see \S\ref{subsec:exp_train_for_orig_zeshel}),
and the following two DE models trained via distillation from the CE.
 \begin{itemize}
    \item \scratchDualEncoder:  DE initialized with 
    \bert~\cite{devlin-etal-2019-bert} and trained \emph{only} using
    training signal from the cross-encoder model.
    \item \finetuneDualEncoder: \fixedDualEncoder model 
    further fine-tuned via distillation using the cross-encoder model.
\end{itemize}
We refer the reader to Appendix~\ref{subsec:appendix_train_de_for_ce} for 
hyper-parameter and optimization details.

\paragraph{Evaluation metric}
We evaluate all approaches under the following two settings.
\begin{itemize}[topsep=1pt,itemsep=0ex,partopsep=1ex,parsep=1ex]
    \item In the first setting, we retrieve $\nRetrievedItems$ items for a given query, re-rank 
            them using exact CE scores and keep top-$k$ items.
            We evaluate each method using Top-$k$-Recall@$\nRetrievedItems$ which is the percentage
            of top-$k$ items according to the CE model present
            in the $\nRetrievedItems$ retrieved items. 
    \item In the second setting, we operate under a fixed test-time cost budget where the cost
            is defined as the number of CE calls made during inference.
            Baselines such as DE and \tfidf will use 
            the entire cost budget for re-ranking
            items using exact CE scores while our proposed approach
            will have to split the budget
            between the number of anchor items ($\nAnchorItems$) 
            used for embedding the query~(\S\ref{subsec:inference}) 
            and the number of items ($\nRetrievedItems$)
            retrieved for final re-ranking. 
\end{itemize}

We refer to our proposed method as~\textbf{$\propMethod_{\nAnchorItems}$}
when using fixed set of $\nAnchorItems$ anchor items chosen
uniformly at random, and
we refer to it as \textbf{\propMethod} when operating under a fixed test-time cost budget in which
case different values of $\nAnchorItems$ and $\nRetrievedItems$ are used in each setting.

\subsubsection{Results}
\begin{figure*}
    \centering
    \includegraphics[width=\textwidth]{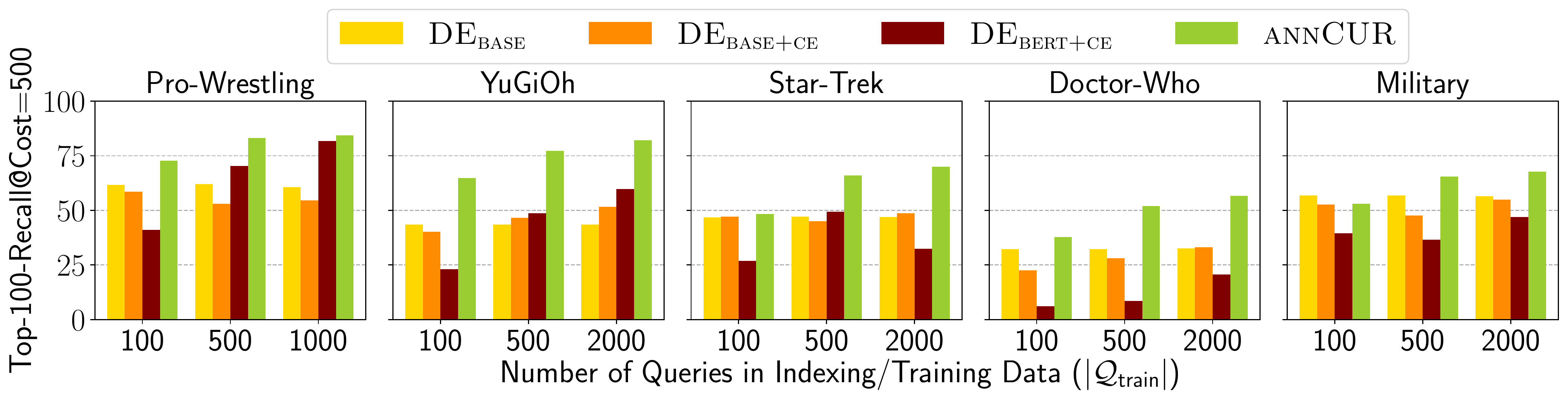}
    \caption{Bar plot showing Top-100-Recall@Cost=500 for different methods as we increase $\queryTrainSize$, the size of 
    indexing/training data for five different domains. }
    \label{fig:train_size_vs_performance_comparison_zeshel}
    \vspace{-0.5cm}
\end{figure*}

Figures~\ref{fig:recall_at_same_k_retrieved_yugioh} and~\ref{fig:recall_at_same_cost_yugioh}  
show recall of top-$k$ cross-encoder nearest neighbors for $k \in \{1, 10, 50, 100\}$
on \zeshel domain = \yugioh when using 500 queries for training,
and evaluating on the remaining 2874 test queries.
Figure~\ref{fig:recall_at_same_k_retrieved_yugioh} shows recall 
when each method retrieves the same number of items and 
Figure~\ref{fig:recall_at_same_cost_yugioh} shows recall 
when each method operates under a fixed inference cost budget.

\paragraph{Performance for $k \geq 10$} 
In Figure~\ref{fig:recall_at_same_k_retrieved_yugioh}, our proposed approach outperforms all baselines 
at finding top-$k=10,50$, and $100$ nearest neighbors 
when all models retrieve the same number of items for re-ranking.
In Figure~\ref{fig:recall_at_same_cost_yugioh}, when operating under the same cost budget, 
\propMethod outperforms DE baselines at larger cost budgets for $k=10,50$, and $100$.
Recall that at a smaller cost budget, \propMethod is able to retrieve 
fewer number of items for exact re-ranking than the baselines as it needs to 
use a fraction of the cost budget i.e. CE calls to compare the test-query
with anchor items in order to embed the query for retrieving relevant items.
Generally, the optimal budget split between the number of anchor items ($\nAnchorItems$) 
and the number of items retrieved for exact re-ranking ($\nRetrievedItems$)
allocates around 40-60\% of the budget to $\nAnchorItems$ and the remaining
budget to $\nRetrievedItems$.

\paragraph{Performance for $k=1$} 
The top-$1$ nearest neighbor according to the CE is likely to be the ground-truth entity (item)
for the given mention (query).
Note that \fixedDualEncoder was trained using a massive amount of entity linking data (all eight training domains in \zeshel, see~\S\ref{subsec:exp_train_for_orig_zeshel}) using the ground-truth entity (item)
as the positive item. 
Thus, it is natural for top-$1$ nearest neighbor 
for both of these models to be aligned. 
For this reason, we observe that \fixedDualEncoder and \finetuneDualEncoder 
outperform \propMethod for $k=1$.
However, our proposed approach either outperforms or is competitive with \scratchDualEncoder,
a DE model trained \emph{only} using CE scores for 500 queries after initializing with \bert.
In Figure~\ref{fig:recall_at_same_k_retrieved_yugioh}, $\propMethod_{100}$ and $\propMethod_{200}$ 
outperform \scratchDualEncoder and in Figure~\ref{fig:recall_at_same_cost_yugioh}
\propMethod outperforms \scratchDualEncoder at larger cost budgets.

We refer the reader to Appendix~\ref{subsec:appendix_exhaustive_exp_res} for results
on all combinations of top-$k$ values, domains, and training data size values.

\paragraph{Effect of training data size ($\queryTrainSize$)}
Figure~\ref{fig:train_size_vs_performance_comparison_zeshel} shows Top-100-Recall@Cost=500
on test queries for various methods as we increase the number of queries in training data~($\queryTrainSize$). 
For DE baselines, the trend is not consistent across all domains. 
On \yugioh, the performance consistently improves with $\queryTrainSize$.
However, on \military, the performance of distilled DE drops on going from 
100 to 500 training queries but improves on going from 500 to 2000 training queries. 
Similarly, on \proWrestling, performance of distilled 
\finetuneDualEncoder does not consistently
improve with training data size while it does for \scratchDualEncoder.
We suspect that this is due to a combination of various factors such as 
overfitting on training data, sub-optimal hyper-parameter configuration, divergence
of model parameters etc.
In contrast, our proposed method, \propMethod, always shows consistent improvements as 
we increase the number of queries in training data, and avoids the perils of gradient-based training
that often require large amounts of training data to avoid overfitting as well expensive 
hyper-parameter tuning in order to consistently work well across various domains.

\begin{figure}
    \centering
    \includegraphics[width=0.48\textwidth]{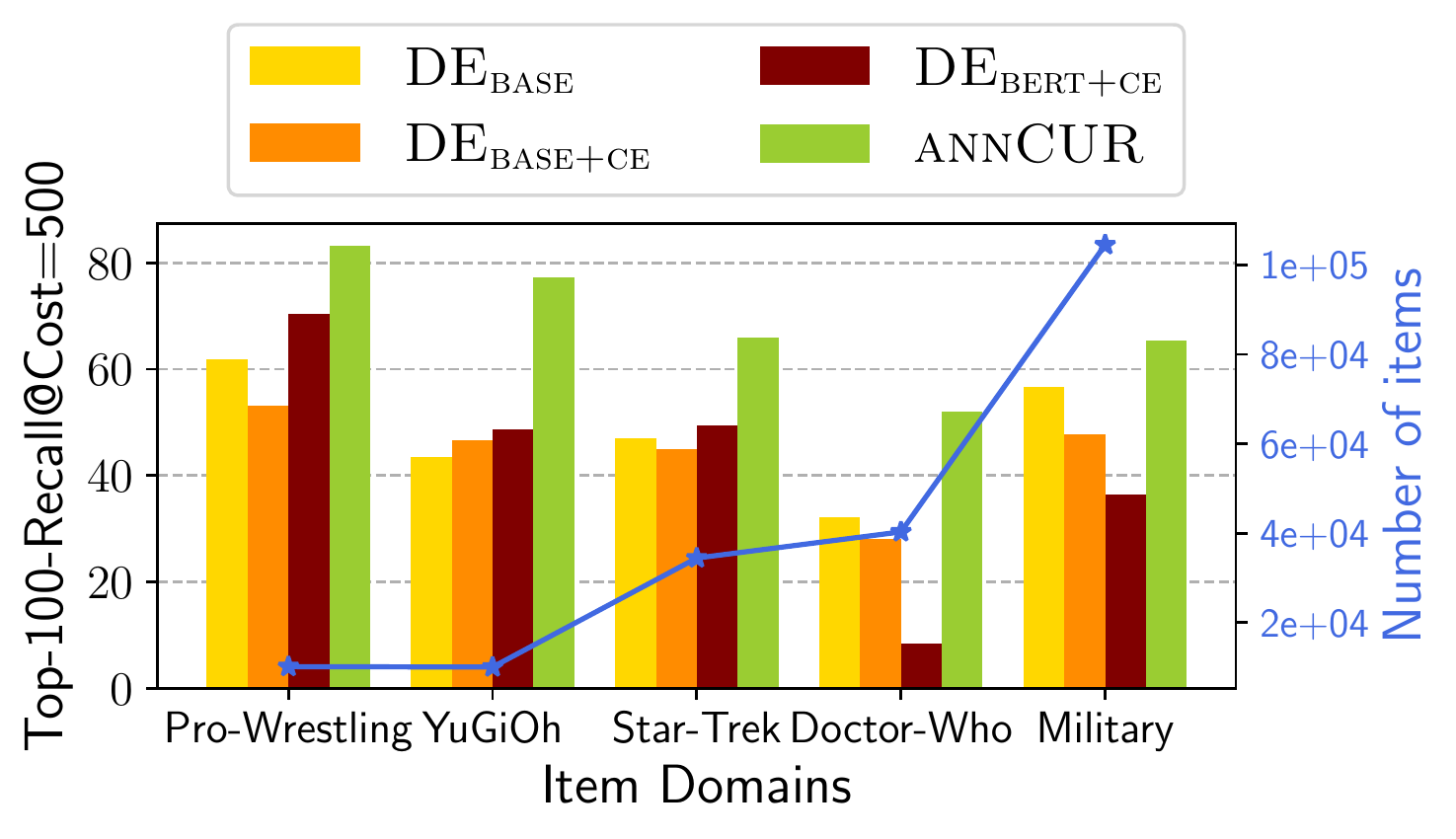}
    \caption{Bar plot with Top-100-Recall@Cost=500 for five domains in \zeshel when using $\queryTrainSize=500$ queries for training/indexing 
    and a line-plot showing the number of items (entities) in each domain. 
    }
    \label{fig:domain_size_vs_performance_comparison_zeshel}
    \vspace{-0.5cm}
\end{figure}

\paragraph{Effect of domain size}
Figure~\ref{fig:domain_size_vs_performance_comparison_zeshel} shows Top-100-Recall@Cost=500
for \propMethod and DE baselines on primary y-axis and 
size of the domain i.e. total number of items on secondary y-axis 
for five different domains in \zeshel. 
Generally, as the number of items in the domain increases, 
the performance of all methods drops.

\paragraph{Indexing Cost }
The indexing process starts by computing query-item CE scores for queries in train split.
\propMethod uses these scores for indexing the items~(see \S\ref{subsec:indexing}) while 
DE baselines use these scores to find ground-truth top-$k$ items for each query 
followed by training DE models using CE query-item scores.
For domain=\yugioh with $10031$ items, and $\queryTrainSize=500$, the time taken to 
compute query-item scores for train/anchor queries ($\queryItemCompTime$) $\approx 10$ 
hours on an NVIDIA GeForce RTX2080Ti GPU/12GB memory,
and training a DE model further takes additional time $(\dualEncTrainTime) \approx$ 
4.5 hours on two instances of the same GPU.
Both $\queryItemCompTime$ and $\dualEncTrainTime$ increase linearly with 
domain size and $\queryTrainSize$, however the query-item score computation can be trivially parallelized.
We ignore the time to build a nearest-neighbor search index over item embeddings produced by
\propMethod or DE as that is negligible in comparison to time spent on CE score 
computation and training of DE models.
We refer the reader to Appendix~\ref{subsec:appendix_train_de_for_ce} for more details.

\subsection{Analysis of \propMethod}
\label{subsec:cur_analysis}

\begin{figure}
     \centering
     \begin{subfigure}[b]{0.23\textwidth}
         \centering
         \includegraphics[width=\textwidth]{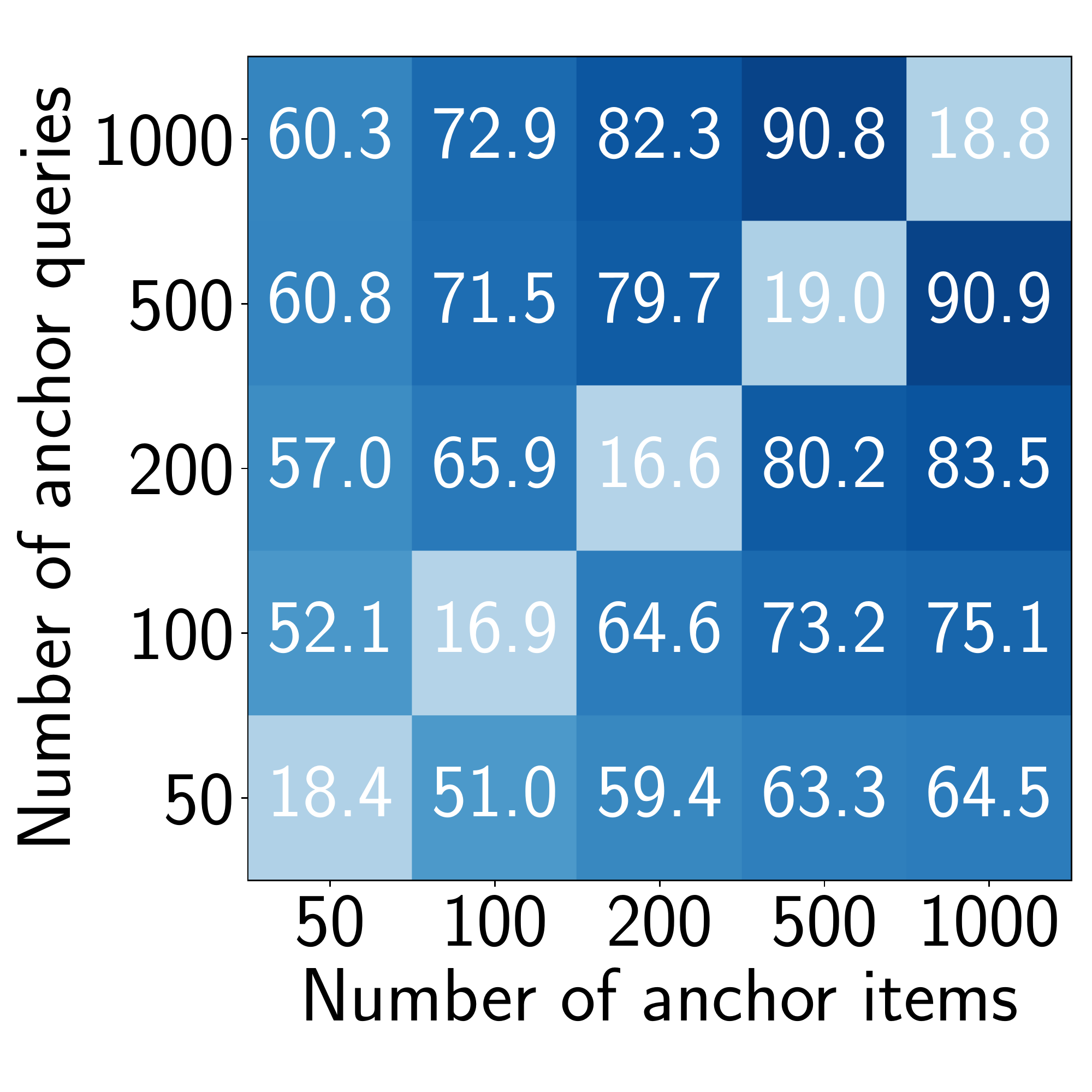}
         \caption{\clsCrossenc}
         \label{fig:arch_change_yugioh_prec_at_k_heatmap_clscrossenc}
     \end{subfigure}
     \begin{subfigure}[b]{0.23\textwidth}
         \centering
         \includegraphics[width=\textwidth]{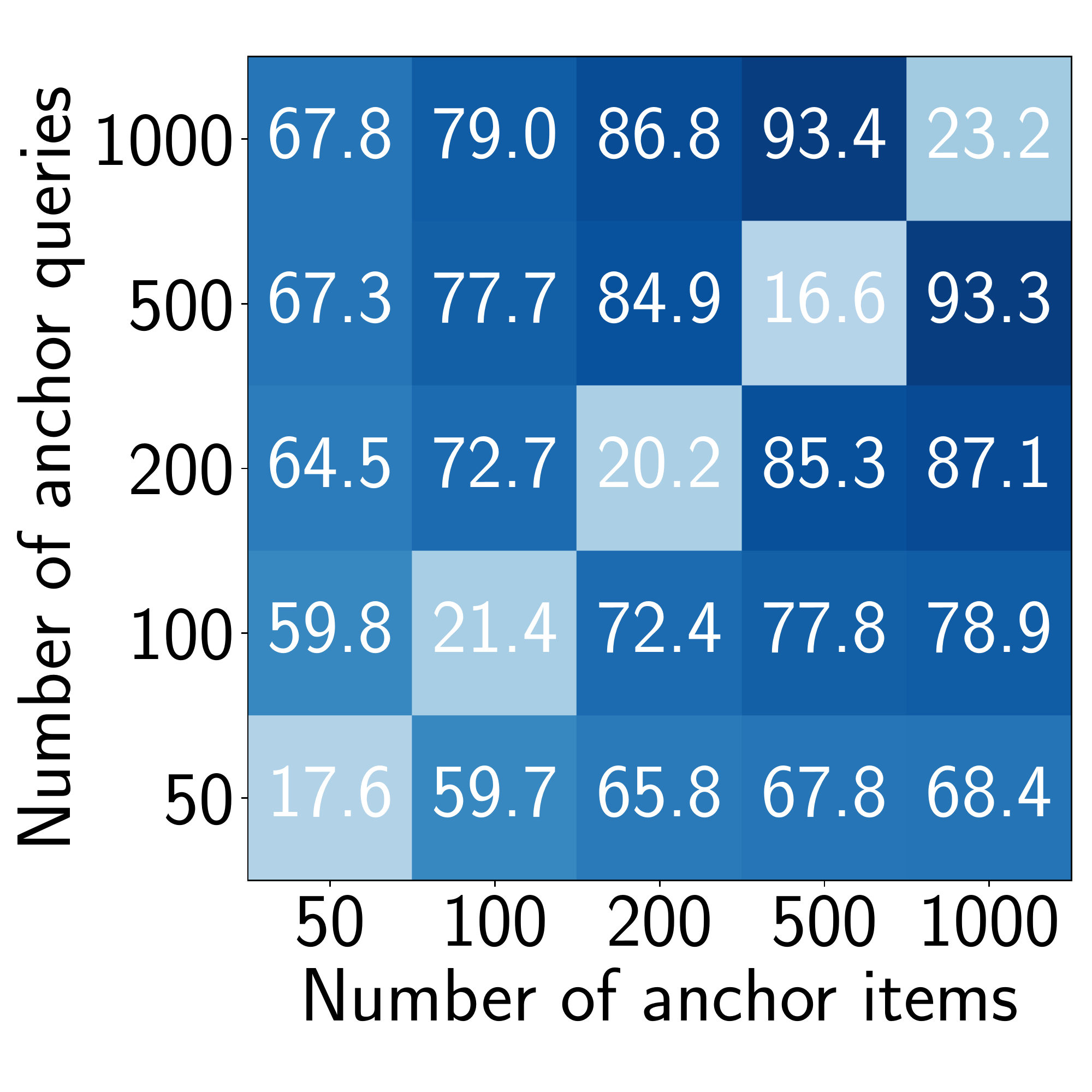}
         \caption{\eCrossenc}
         \label{fig:arch_change_yugioh_prec_at_k_heatmap_ecrossenc}
     \end{subfigure}
    \caption{Top-10-Recall@500 of \propMethod for non-anchor queries on domain = \texttt{YuGiOh} for two cross-encoder models -- \clsCrossenc and \eCrossenc.}
    \label{fig:arch_change_yugioh_prec_at_k_heatmap}
    \vspace{-0.5cm}
\end{figure}

We compute the query-item score matrix for both \clsCrossenc and \eCrossenc and
compute the rank of these matrices using \texttt{numpy}~\cite{harris2020array} for domain=\yugioh
with 3374 queries (mentions) and 10031 items (entities).
Rank of the score matrix for \clsCrossenc = 315 which is much higher than rank of
the corresponding matrix for \eCrossenc = 45 due to the query-item score 
distribution produced by \clsCrossenc model being much more skewed than 
that produced by \eCrossenc model 
(see Fig.~\ref{fig:score_dist_all_models}).

Figures~\ref{fig:arch_change_yugioh_prec_at_k_heatmap_clscrossenc} 
and~\ref{fig:arch_change_yugioh_prec_at_k_heatmap_ecrossenc} show
Top-10-Recall@500 on domain=\yugioh  for \clsCrossenc and \eCrossenc respectively
on different combinations of number of anchor 
queries ($\nAnchorQueries$) and anchor items ($\nAnchorItems$). 
Both anchor queries and anchor items are chosen uniformly at random,
and for a given set of anchor queries, we evaluate on the 
remaining set of queries.

\paragraph{\clsCrossenc versus \eCrossenc}
For the same choice of anchor queries and anchor items, 
the proposed method performs better with \eCrossenc model 
as compared \clsCrossenc due to the query-item
score matrix for \eCrossenc having much lower rank
thus making it easier to approximate.

\paragraph{Effect of $\nAnchorQueries$ and $\nAnchorItems$}
Recall that the indexing time for \propMethod is directly proportional to
the number of anchor queries ($\nAnchorQueries$) while the
number of anchor items ($\nAnchorItems$) influences the 
test-time inference latency.
Unsurprisingly, performance of \propMethod increases as we increase 
$\nAnchorItems$ and $\nAnchorQueries$, and these can be tuned as per user's 
requirement to obtain desired recall-vs-indexing time and recall-vs-inference time trade-offs.
We refer the reader to Appendix~\ref{subsec:appendix_equal_anchors_explanation} for a detailed
explanation for the drop in performance when $\nAnchorQueries = \nAnchorItems$.

\subsection{Item-Item Similarity Baselines}
\label{subsec:item_ce_baselines}

\begin{figure}
    \centering
    \includegraphics[width=0.47\textwidth]{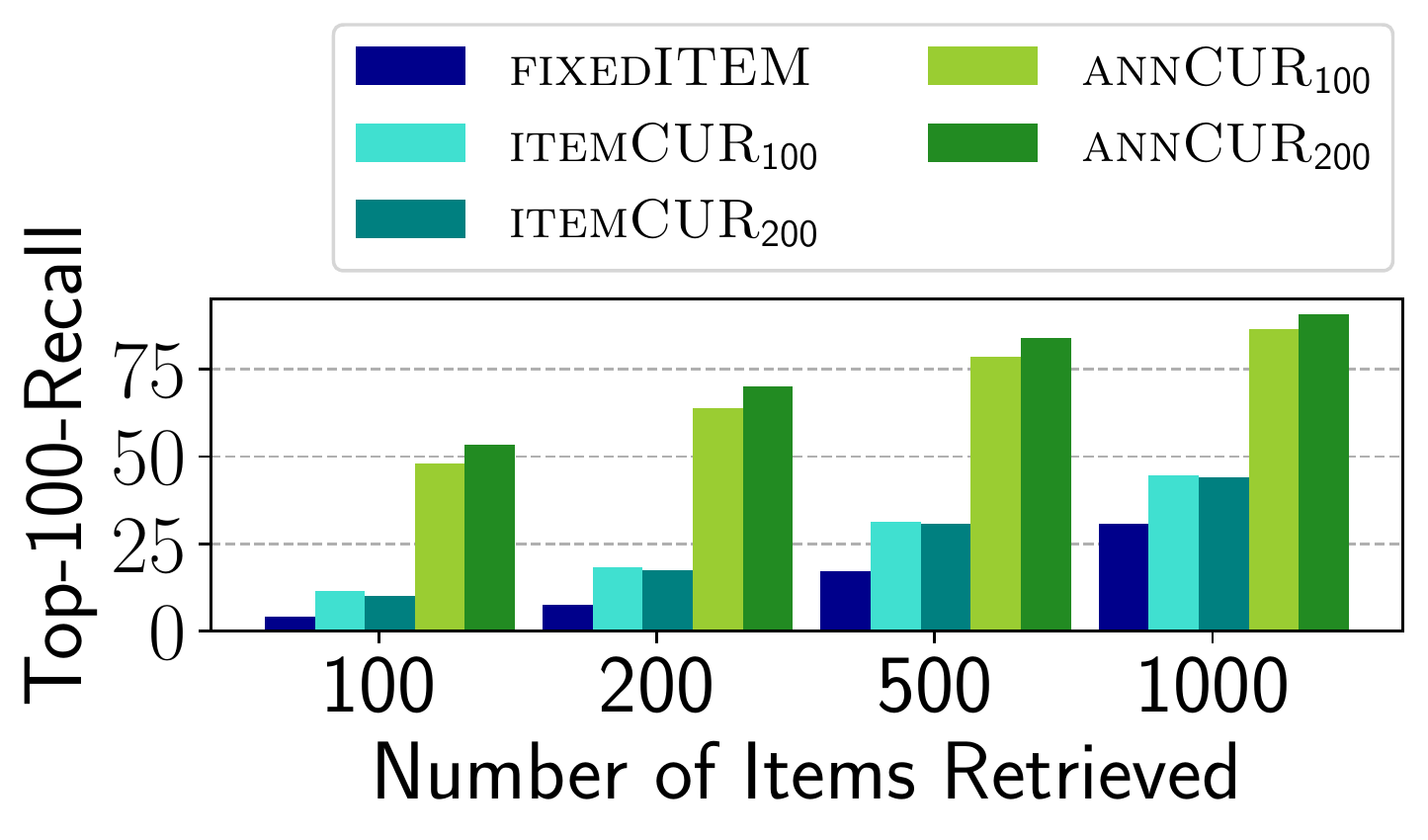}
    \caption{Bar plot showing Top-100-Recall for domain=\yugioh when 
    indexing using 500 anchor items for \fixedItems and \itemCUR and 500 anchor queries for \propMethod. 
    }
    \label{fig:cur_ablation_yugioh}
    \vspace{-0.5cm}
\end{figure}

We additionally compare with the following baselines that 
index items by comparing against a fixed set of anchor 
items~\footnote{See appendix~\ref{subsec:appendix_item_item_ce} for details on 
computing item-item scores using a CE model trained to score query-item pairs.
}
instead of anchor queries.
\begin{itemize}[topsep=1pt,itemsep=0ex,partopsep=1ex,parsep=1ex]
    \item \fixedItems: Embed all items and test-query in $\RR^{\nAnchorItems}$ 
    using CE scores with a fixed set of $\nAnchorItems$ items chosen uniformly
    at random, and retrieve top-$\nRetrievedItems$ items for the
    test query based on dot-product of these $\nAnchorItems$-dim embeddings. We use $\nAnchorItems=500$.
    \item \itemCUR-$\nAnchorItems$: This is similar to the proposed approach except that it 
    indexes the items by comparing them against $\nAnchorItemsForIndex$ anchor 
    items instead of anchor queries
    for computing $\rowMatrixTrain$ and $\colMatrixTrain$ matrices 
    in the indexing step in~\S\ref{subsec:indexing}. At test time, 
    it performs inference just like \propMethod~(see \S\ref{subsec:inference}) 
    by comparing against a different set of fixed $\nAnchorItems$ anchor items.
    We use $\nAnchorItemsForIndex=500$.
\end{itemize}

Figure~\ref{fig:cur_ablation_yugioh} shows Top-100-Recall for 
\fixedItems, \itemCUR, and \propMethod on domain = \yugioh.
\itemCUR performs better than \fixedItems indicating
that the latent item embeddings produced using CUR decomposition 
of the item-item similarity matrix are better than those built 
by comparing the items against a fixed set of anchor items.
\itemCUR performs worse than \propMethod apparently because the CE
was trained on query-item pairs and was not calibrated for item-item comparisons.

\section{Related Work}
\label{sec:related_work}
\paragraph{Matrix Decomposition}
Classic matrix decomposition methods such as SVD, QR decomposition 
have been used for approximating kernel matrices
and distance matrices~\cite{musco2017sublinear, tropp2017randomized, bakshi2018sublinear, indyk2019sample}.
Interpolative decomposition methods such as \nystrom method and CUR decomposition
allow approximation of the matrix even when given only a subset of rows and columns of 
the matrix. Unsurprisingly, performance of these methods
can be further improved if given the entire matrix as it allows 
for a better selection of rows and columns on the matrix used in the decomposition 
process~\cite{goreinov1997theory, drineas2005nystrom, kumar2012sampling, wang2013improving}.
Recent work, \citet{archan2022sublinear} proposes sublinear \nystrom approximations and considers CUR-based approaches for approximating non-PSD similarity matrices that arise in NLP tasks such as coreference resolution and document classification.
Unlike previous work, our goal is to use the approximate scores to support
retrieval of top scoring items. 
Although matrix decomposition methods for sparse matrices 
based on SVD~\cite{berry1992large, keshavan2010matrix, hastie2015matrix, ramlatchan2018survey} 
can be used instead of CUR decomposition, such methods would require 
a) factorizing a sparse matrix at \emph{test time} in order to obtain latent embeddings for all items and the test query,
and b) indexing the latent item embeddings to efficiently retrieve top-scoring items for the given query.
In this work, we use CUR decomposition as, unlike other sparse matrix decomposition methods,
CUR decomposition allows for \emph{offline} computation and indexing of item embeddings and 
the latent embedding for a test query is obtained simply by using its cross-encoder scores against the anchor items.

\paragraph{Cross-Encoders and Distillation}
Due to high computational costs, use of cross-encoders (CE) is often limited
to either scoring a fixed set of items or re-ranking items retrieved 
by a separate (cheaper) retrieval model~\cite{logeswaran-etal-2019-zero,  qu-etal-2021-rocketqa,bhattacharyya-etal-2021-energy, ayoola2022refined}.
CE models are also widely used for training 
computationally cheaper models via distillation on the training domain~\citep{wu-etal-2020-scalable,reddi2021rankdistil},
or for improving performance of these cheaper models on the target domain~\cite{chen-etal-2020-dipair, thakur-etal-2021-augmented}
by using cross-encoders to score a fixed or heuristically retrieved set of items/datapoints.
The DE baselines used in this work, in contrast, are trained 
using $k$-nearest neighbors for a given query according to the CE.

\paragraph{Nearest Neighbor Search}
For applications where the inputs are described as vectors in $\RR^{n}$,
nearest neighbor search has been widely studied for various  (dis-)similarity functions
such as $\ell_2$ distance~\cite{chavez2001searching, hjaltason2003index}, inner-product~\cite{jegou2010product, johnson2019billion, guo2020accelerating}, and Bregman-divergences~\cite{cayton2008fast}.
Recent work on nearest neighbor search with non-metric (parametric)
similarity functions explores various tree-based~\cite{boytsov2019pruning} 
and graph-based nearest neighbor search indices~\cite{boytsov2019accurate, tan2020fast, tan2021fast}.
In contrast, our approach approximates the scores
of the parametric similarity function using the latent embeddings generated
using CUR decomposition and uses off-the-shelf maximum inner product search methods 
with these latent embeddings to find $k$-nearest neighbors for the CE. 
An interesting avenue for future work would be 
to combine our approach with tree-based and graph-based approaches 
to further improve efficiency of these search methods.

\section{Conclusion}

In this paper, we proposed, \propMethod, a matrix factorization-based
approach for nearest neighbor search for a cross-encoder model
without relying on an auxiliary model such as a dual-encoder for retrieval.
\propMethod approximates the test query's scores with all items by 
scoring the test-query only with a small number of anchor items,
and retrieves items using the approximate scores.
Empirically, for $k > 10$, our approach provides test-time 
recall-vs-computational cost trade-offs 
superior to the widely-used approach of using cross-encoders to re-rank items 
retrieved using a dual-encoder or a \tfidf-based model. This work is a step
towards enabling efficient retrieval with expensive similarity functions
such as cross-encoders, and thus, moving beyond using 
such models merely for re-ranking items retrieved by 
auxiliary retrieval models such as dual-encoders and \tfidf-based models.

\section*{Acknowledgements}
We thank members of UMass IESL for helpful discussions
and feedback. 
This work was supported 
in part by the Center for Data Science and the Center
for Intelligent Information Retrieval, 
in part by the
National Science Foundation under Grant No. NSF1763618, 
in part by the Chan Zuckerberg Initiative under the project “Scientific Knowledge Base Construction”, 
in part by International Business Machines Corporation Cognitive Horizons Network agreement number W1668553, and 
in part using high performance computing equipment obtained
under a grant from the Collaborative R\&D Fund
managed by the Massachusetts Technology Collaborative. 
Rico Angell was supported by the NSF Graduate Research Fellowship
under Grant No. 1938059. 
Any opinions, findings and conclusions or recommendations expressed in this material are
those of the authors and do not necessarily reflect those of
the sponsor(s).

\section*{Limitations}

In this work, we use cross-encoders parameterized using transformer models. 
Computing query-item scores using such models can be computationally expensive.
For instance, on an NVIDIA GeForce RTX 2080Ti GPU with 12GB memory, we
can achieve a throughput of approximately 140 scores/second, and
computing a score matrix for 100 queries and 10K items takes about two hours.
Although this computation can be trivially parallelized, the total amount of GPU hours
required for this computation can be very high.
However, note that these scores need to be computed even for distillation
based DE baselines as we need to identify $k$-nearest neighbors for each query
according to the cross-encoder model for training a dual-encoder model on this task.

Our proposed approach allows for indexing the item set only 
using scores from cross-encoder without any additional
gradient based training but it is not immediately
clear how it can benefit from data on multiple target domains at the same time.
Parametric models such as dual-encoders on the other hand can 
benefit from training and knowledge distillation on multiple domains
at the same time.

\section*{Ethical Consideration}

Our proposed approach considers how to speed up the computation of nearest neighbor search for cross-encoder models. The cross-encoder model, which our approach approximates, may have certain biases / error tendencies. Our proposed approach does not attempt to mitigate those biases. It is not clear how those biases would propagate in our approximation, which we leave for future work. An informed user would scrutinize both the cross-encoder model and the resulting approximations used in this work.

\bibliography{main}
\balance
\bibliographystyle{acl_natbib}

\appendix
\section{Training Details}
\label{sec:appendix_training_details}

\subsection{Training DE and CE for Entity Linking on \zeshel}
\label{subsec:appendix_train_de_and_ce_zeshel}

We initialize all models with \texttt{bert-base-uncased} and 
train using Adam~\cite{DBLP:journals/corr/KingmaB14} optimizer with learning rate = $10^{-5}$, 
and warm-up proportion=0.01 for four epochs. We evaluate on dev 
set five times during each epoch, and pick the model checkpoint that
maximises accuracy on dev set.
While training the dual-encoder model, we update negatives after each epoch
using the latest dual-encoder model parameters to mine hard negatives.
We trained the cross-encoder with a fixed set of 63 negatives items (entities) for
each query (mention) mined using the dual-encoder model.
We use batch size of 8 and 4 for training the dual-encoder and cross-encoder respectively.

Dual-encoder and cross-encoder models took 34 an 44 hours respectively for training  
on two NVIDIA GeForce RTX 8000 GPUs each with 48GB memory.
The dual-encoder model has 2$\times$110M parameters as it consists of separate query and item encoder models
while the cross-encoder model has 110M parameters.

\paragraph{Tokenization details}

We use word-piece tokenization~\cite{wu2016google} for with a 
maximum of 128 tokens including special tokens for tokenizing entities and mentions.
The mention representation consists of 
the word-piece tokens of the context surrounding the mention and the mention itself as
\[ \texttt{[CLS] ctxt\textsubscript{l} [M\textsubscript{s}] ment [M\textsubscript{e}] ctxt\textsubscript{r} [SEP]} \]
where \texttt{ment}, \texttt{ctxt\textsubscript{l}}, and \texttt{ctxt\textsubscript{r}} are word-piece
tokens of the mention, context before and after the
mention respectively, and \texttt{[M\textsubscript{s}]}, \texttt{[M\textsubscript{e}]} are special
tokens to tag the mention. 

The entity representation is also composed of word-piece tokens of the entity title
and description. The input to our entity model is:
\[ \texttt{[CLS] title [ENT] description [SEP]} \]
where \texttt{title}, \texttt{description} are word-pieces tokens of
entity title and description, and \texttt{[ENT]} is a special token to separate entity title and description
representation. 

\begin{table}[!ht]
    \small
    \centering
    \begin{tabular}{l|r r c }
        \toprule
        Domain                  &  $\lvert \mathcal{E} \rvert$ & $\lvert \mathcal{M} \rvert$ & $\lvert \mathcal{M}_{k-\text{NN}} \rvert$\\
        \midrule
        \multicolumn{4}{c}{Training Data} \\
        \midrule
        Military                &   104520  & 13063     & 2400 \\
        Pro Wrestling           &   10133   & 1392      & 1392 \\
        Doctor Who              &   40281   & 8334      & 4000 \\
        American Football       &   31929   & 3898      & - \\
        Fallout                 &   16992   & 3286      & - \\
        Star Wars               &   87056   & 11824     & - \\
        World of Warcraft       &   27677   & 1437      & - \\
        \midrule
        \multicolumn{4}{c}{Validation Data} \\
        \midrule
        Coronation Street       &   17809   & 1464      & - \\
        Muppets                 &   21344   & 2028      & - \\
        Ice Hockey              &   28684   & 2233      & - \\
        Elder Scrolls           &   21712   & 4275      & - \\
        \midrule
        \multicolumn{4}{c}{Test Data} \\
        \midrule
        Star Trek               &   34430   & 4227      & 4227 \\
        YuGiOh                  &   10031   & 3374      & 3374 \\
        Forgotten Realms        &   15603   & 1200      & - \\
        Lego                    &   10076   & 1199      & - \\
        \bottomrule
    \end{tabular}
    \caption{Statistics on number of entities $\lvert \mathcal{E} \rvert$ (items), total number of mentions $\lvert \mathcal{M} \rvert$ (queries), and number of mentions used in $k$-NN experiments ($\lvert \mathcal{M}_{k-\text{NN}} \rvert$) in \S\ref{subsec:exp_cur_vs_baselines} for each domain in \zeshel dataset.}
    \label{tab:zeshel_stats}
\end{table}
The cross-encoder model takes as input the concatenated
query (mention) and item (entity) representation with the
\texttt{[CLS]} token stripped off the item (entity) tokenization as shown below
\begin{align*}
   &\texttt{[CLS] ctxt\textsubscript{l} [M\textsubscript{s}] ment [M\textsubscript{e}] ctxt\textsubscript{r} [SEP]} \\
   &\texttt{ title [ENT] description [SEP]} 
\end{align*}

\subsection{Using query-item CE model for computing item-item similarity}
\label{subsec:appendix_item_item_ce}
We compute item-item similarity using a cross-encoder trained to score query-item pairs as follows. 
The query and item in our case correspond to mention of an entity with surrounding context and entity with 
its associated title and description respectively.
We feed in first entity in the pair in the query slot by using mention span tokens around the title of the entity,
and using entity description to fill in the right context of the mention. We feed in the second entity in 
the entity slot as usual. The concatenated representation of the entity pair ($e_1, e_2$) is given by
\[ \small{ \texttt{[CLS] [M\textsubscript{s}] t\textsubscript{e\textsubscript{1}} [M\textsubscript{e}] d\textsubscript{e\textsubscript{1}} [SEP]  t\textsubscript{e\textsubscript{2}} [E] d\textsubscript{e\textsubscript{2}} [SEP]} }\]
where 
\texttt{t\textsubscript{e\textsubscript{1}}, t\textsubscript{e\textsubscript{2}}} are the tokenized titles of the entities,
\texttt{d\textsubscript{e\textsubscript{1}}, d\textsubscript{e\textsubscript{2}}} are the tokenized
entity descriptions,
\texttt{[M\textsubscript{e}]}, \texttt{[M\textsubscript{s}]} are special tokens denoting mention span boundary and \texttt{[E]} is a special token separating entity title from its description.

\subsection{Training DE for $k$-NN retrieval with CE}
\label{subsec:appendix_train_de_for_ce}

We train dual-encoder models using $k$-nearest neighbor 
items according to cross-encoder model for each query using two loss functions.
Let $\scoreMat{DE}$ and $\scoreMat{CE}$ be matrices 
containing score for all items for each query in training data.
Let $\topKEnts{\kDistill}{CE}{\ment}, \topKEnts{\kDistill}{DE}{\ment}$ be top-$\kDistill$ items
for query $\ment$ according to the cross-encoder
and dual-encoder respectively, and let $\topKNegEnts{\kDistill}{DE}{\ment}$ 
be top-$\kDistill$ items for query $\ment$ according to dual-encoder 
that are not present in $\topKEnts{\kDistill}{CE}{\ment}$.

We use loss functions $\loss{match}$ and $\loss{pair}$ described below 
for training the dual-encoder model using a cross-encoder model.

\[ \loss{match} =  \sum_{\ment \in \queryTrainData}  \crossEntropy \Big(\softmax( \scoreMat{DE}_{[\ment, :]}), \softmax(\scoreMat{CE}_{[\ment, :]} ) \Big) \]
where $\crossEntropy$ is the cross-entropy function, and $\softmax(.)$ is the softmax function.
In words, $\loss{match}$ is the cross-entropy loss between the dual-encoder
and cross-encoder query-item 
score distribution over \emph{all} items.
Due to computational and memory limitations, we train by minimizing 
$\loss{match}$ using items in $\topKEnts{\kDistill}{CE}{\ment}$ for each query $q \in \queryTrainData$.

\[ \loss{pair} = \sum_{\ment \in \queryTrainData} \sum_{(i,j) \in \posNegPairs{\ment} } \crossEntropy([0,1], \softmax([\scoreMat{DE}_{\ment, i}, \scoreMat{DE}_{\ment, j}]) )  \]
\[\text{where}, \hfill \posNegPairs{\ment} = \{ ( \topKEnts{\kDistill}{CE}{\ment}_j, \topKNegEnts{\kDistill}{DE}{\ment}_j ) \}_{j=0}^{\kDistill} \]
$\loss{pair}$ treats items in $\topKEnts{\kDistill}{CE}{\ment}$ 
as a positive item, pairs it with hard negatives
from $\topKNegEnts{\kDistill}{DE}{\ment}$, and minimizing $\loss{pair}$ 
increases dual-encoder's score for items
in $\topKEnts{\kDistill}{CE}{\ment}$, thus aligning $\topKEnts{\kDistill}{DE}{\ment}$ 
with $\topKEnts{\kDistill}{CE}{\ment}$ for queries in training data.

\paragraph{Training and optimization details}
We train all dual-encoder models using Adam optimizer with learning rate=$10^{-5}$
for 10 epochs. We use a separate set of parameters for query and item encoders.
We use 10\% of training queries for validation and train on the remaining
90\% of the queries. For each domain and training data size, we train
with both $\loss{match}$ and $\loss{pair}$ loss functions, and pick the model
that performs best on validation queries for $k$-NN retrieval according to the cross-encoder model.

We train models for with loss $\loss{pair}$ on two NVIDIA GeForce RTX 2080Ti GPUs with 12GB GPU memory
and with loss $\loss{match}$ on two NVIDIA GeForce RTX 8000 GPUs with 48GB GPU memory as we could not train
with $\kDistill=100$ on 2080Tis due to GPU memory limitations.
For loss $\loss{pair}$, we update the the list of negative items ($\topKNegEnts{\kDistill}{DE}{\ment}$ )
for each query after
each epoch by mining hard negative items using the latest dual-encoder model parameters.

\begin{table}[!ht]
    \small
    \begin{subtable}[t]{0.45\textwidth}
        \centering
        \begin{tabular}{c l| c c c c}
        \toprule
        $\queryTrainSize$ &    Model  &  $t_\text{CE-Mat}$  & $t_\text{train}$ & $t_\text{total}$  \\
        \midrule
        100             & DE-$\loss{pair}$      &       2                            &      2.5         &   4.5\\  
        100             & DE-$\loss{match}$     &       2                            &      0.5         &   2.5\\  
        100             & \propMethod           &       2                            &        -         &   2 \\
        \midrule
        500             & DE-$\loss{pair}$      &       10                           &     4.5          &  14.5 \\  
        500             & DE-$\loss{match}$     &       10                           &      1           &  11 \\  
        500             & \propMethod           &       10                           &     -            &  10 \\
        \midrule
        2000             & DE-$\loss{pair}$     &       40                           &      11          &  51 \\  
        2000             & DE-$\loss{match}$    &       40                           &      3           &  43 \\  
        2000            & \propMethod           &       40                           &       -          &  40 \\
        \bottomrule
        \end{tabular}
        \caption{Indexing time (in hrs) for \propMethod and distillation based DE baselines for different 
        number of anchor/train queries ($\queryTrainSize$) for domain=\yugioh.}
        \label{tab:indexing_time_comparison_yugioh}
    \end{subtable}
    \hfill
    \vspace{0.5cm}
    \begin{subtable}[t]{0.45\textwidth}
        \centering
        \begin{tabular}{l c|c c c}
        \toprule
        Domain (w/ size) &    Model  &  $t_\text{CE-Mat}$  & $t_\text{train}$ & $t_\text{total}$ \\
        \midrule
        \yugioh-10K         & DE-$\loss{pair}$      &   10      &   4.5     & 14.5  \\  
        \yugioh-10K         & \propMethod           &   10      &   -       & 10 \\
        \midrule
        \proWrestlingShort-10K   & DE-$\loss{pair}$ &   10      &   4.4     & 14.4  \\  
        \proWrestlingShort-10K   & \propMethod      &   10      &   -       & 10  \\
        \midrule
        \starTrek-34K       & DE-$\loss{pair}$      &   40      &   5.1       & 45.1  \\  
        \starTrek-34K       & \propMethod           &   40      &   -       & 40  \\
        \midrule
        \doctorWho-40K      & DE-$\loss{pair}$      &   40      &   5.2     & 45.2  \\  
        \doctorWho-40K      & \propMethod           &   40      &   -       & 40  \\
        \midrule
        \military-104K       & DE-$\loss{pair}$     &   102     &   5.1       & 107.1 \\  
        \military-104K       & \propMethod          &   102     &   -       & 102  \\
        \bottomrule
        \end{tabular}
        \caption{Indexing time (in hrs) for \propMethod and distillation based DE baselines for various domains
        when using $\queryTrainSize$=500 anchor/train queries.}
        \label{tab:indexing_time_comparison_all_domains}
    
    \end{subtable}
    \caption{Indexing time breakdown for \propMethod and DE models trained via distillation.}
\end{table}

\begin{table*}[]
\small
    \begin{subtable}{0.55\textwidth}
        \centering
        \begin{tabular}{c | c c | c c}
            \toprule
                Training & \multicolumn{2}{c|}{Dev Set}   & \multicolumn{2}{c}{Test Set} \\
               Negatives & \clsCrossenc  & \eCrossenc    &  \clsCrossenc & \eCrossenc \\
            \midrule
            Random              &  59.60        & 57.74         &  58.72        & 56.56     \\
            \tfidf              &  62.19        & 62.29         &  58.20        & 58.36     \\
            DE                  &  67.67        & 66.86         &  65.87        & 65.49     \\
            \bottomrule
        \end{tabular}
        \caption{Macro-Average of Entity Linking Accuracy for \clsCrossenc and \eCrossenc 
        models on test and dev set in \zeshel.}
        \label{tab:arch_change_accuracy_comparison}
    \end{subtable}
    \hfill
    \begin{subtable}{0.42\textwidth}
    \centering
    \begin{tabular}{c|c c}
        \toprule
        \multirow{1}{*}{Training}  &  \multirow{2}{*}{\clsCrossenc} & \multirow{2}{*}{\eCrossenc} \\
        Negatives\\
        \midrule
        Random              &  816          & 354   \\
        \tfidf              &  396          & 67    \\
        DE                  &  315          & 45    \\
        \bottomrule
    \end{tabular}
    \caption{Rank of $3374 \times 10031$ mention-entity cross-encoder score matrix for test domain = \yugioh }
    \label{tab:arch_change_rank_yugioh}
    \end{subtable}
    \caption{Accuracy on the downstream task of entity linking and rank of query-item (mention-entity) score matrix
    for \clsCrossenc and \eCrossenc trained using different types of negatives.}
\end{table*}

\paragraph{Indexing and Training Time}
Table~\ref{tab:indexing_time_comparison_yugioh} shows overall indexing time for the proposed method \propMethod
and for DE models trained using two distillation losses -- $\loss{pair}$ and $\loss{match}$ on domain=\yugioh.
Training time ($t_\text{train}$) for loss $\loss{match}$ is much less as compared to that for $\loss{pair}$ as the 
former is trained on more powerful GPUs (two NVIDIA RTX8000s with 48GB memory each) 
due to its GPU memory requirements while the latter is trained
on two NVIDIA 2080Ti GPUs with 12 GB memory each.
The total indexing time ($t_\text{total})$ for DE models includes the time
taken to compute CE score matrix ($t_\text{CE-Mat}$) because 
in order to train a DE model for the task of $k$-nearest neighbor
search for a CE, we need to first find exact $k$-nearest neighbor items for the
training queries. 
Note that this is different from the "standard" way of training of DE models via distillation
where the DE is often distilled using CE scores on a \emph{fixed or heuristically retrieved} set of items,
and \emph{not} on $k$-nearest neighbor items according to the cross-encoder for a given query.

Table~\ref{tab:indexing_time_comparison_all_domains} shows indexing time for \propMethod and DEs 
trained via distillation for five domains in \zeshel. As the size of the domain increases,
the time take for computing cross-encoder scores on training queries ($t_\text{CE-Mat}$) also increases.
The time takes to train dual-encoder via distillation roughly remains the same as we train with
fixed number of positive and negative items during distillation.

\section{Additional Results and Analysis}
\subsection{Comparing \eCrossenc and \clsCrossenc}
\label{subsec:appendix_cls_vs_ecrossenc}

In addition to training cross-encoder models with negatives mined using
a dual-encoder, we train both \clsCrossenc and \eCrossenc models using 
random negatives and negatives mined using \tfidf embeddings of mentions and entities. To evaluate the cross-encoder models, we retrieve 64 entities for each 
test mention using a dual-encoder model and re-rank them using a cross-encoder model. 

Table~\ref{tab:arch_change_accuracy_comparison} shows macro-averaged accuracy 
on the downstream task of entity linking
over test and dev domains in \zeshel dataset,
and Table~\ref{tab:arch_change_rank_yugioh} shows rank of query-item score matrices
on domain=\yugioh for both cross-encoder models.
The proposed \eCrossenc model performs at par with the widely used \clsCrossenc
architecture for all three kinds of negative mining strategies while producing a
query-item score matrix with lower rank as compared to \clsCrossenc.

\begin{figure}[!h]
     \centering
     \begin{subfigure}[b]{0.23\textwidth}
         \centering
         \includegraphics[width=\textwidth]{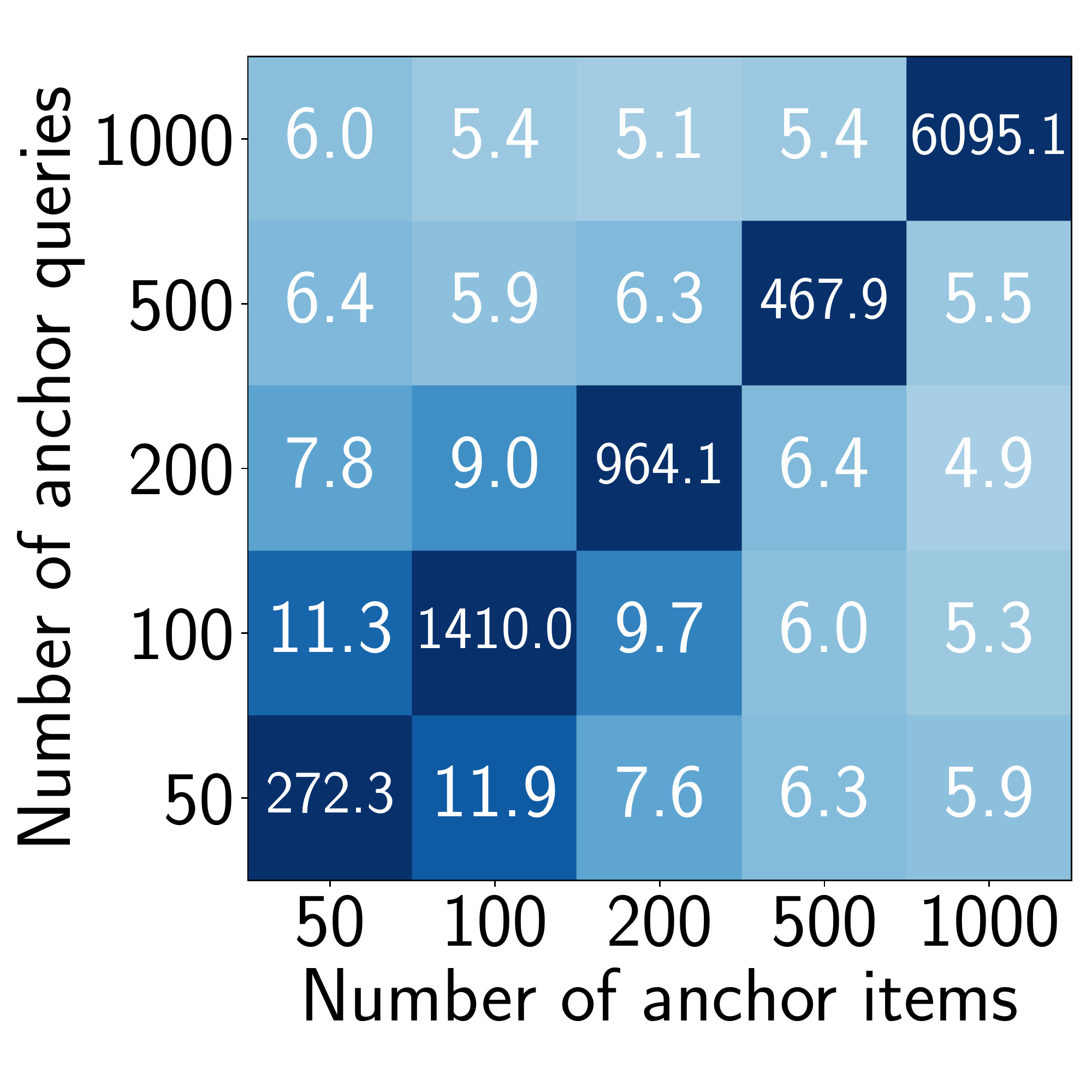}
         \caption{\clsCrossenc}
         \label{fig:arch_change_yugioh_approx_error_heatmap_clscrossenc}
     \end{subfigure}
     \hfill
     \begin{subfigure}[b]{0.23\textwidth}
         \centering
         \includegraphics[width=\textwidth]{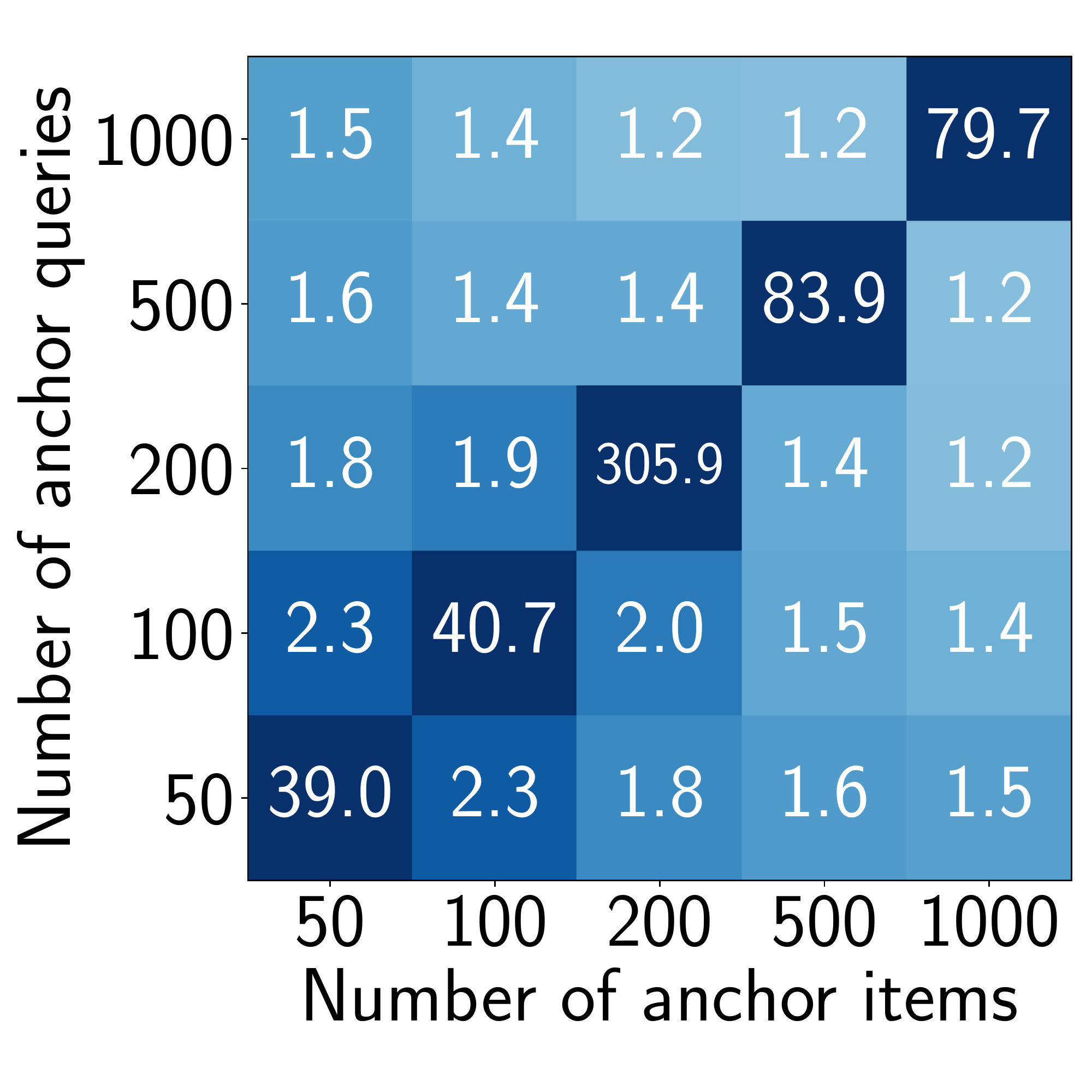}
         \caption{\eCrossenc}
         \label{fig:arch_change_yugioh_approx_error_heatmap_ecrossenc}
     \end{subfigure}
    \caption{
    Matrix approximation error evaluated on non-anchor queries for CUR decomposition on domain =  \yugioh 
    for \clsCrossenc and \eCrossenc models. The total number of queries including both anchor and non-anchor (test)
    queries is 3374 and the total number of items is 10031.}
    \label{fig:arch_change_yugioh_approx_error_heatmap}
\end{figure}

Figure~\ref{fig:arch_change_yugioh_approx_error_heatmap} shows
approximation error of \propMethod for different
combinations of number of anchor queries and anchor items for \clsCrossenc and \eCrossenc. 
For a given set of anchor queries, the approximation error is evaluated
on the remaining set of queries.
The error between a matrix $M$ and its approximation $\Tilde{M}$ is measured as $\|M~-~\Tilde{M}\|_{F}/~\|M\|_F$
where $\|.\|_F$ is the Frobenius norm of a matrix.
For the same choice of anchor queries and anchor items, the approximation
error is lower for \eCrossenc model as compared to \clsCrossenc. This aligns
with the observation that rank of the query-item score matrix from \eCrossenc is lower
than the corresponding matrix from \clsCrossenc as shown in Table~\ref{tab:arch_change_rank_yugioh}.

\subsection{Understanding poor performance of \propMethod for $\nAnchorItems = \nAnchorQueries$}
\label{subsec:appendix_equal_anchors_explanation}

\begin{figure}[!ht]
     \centering
     \begin{subfigure}[b]{0.23\textwidth}
         \centering
         \includegraphics[width=\textwidth]{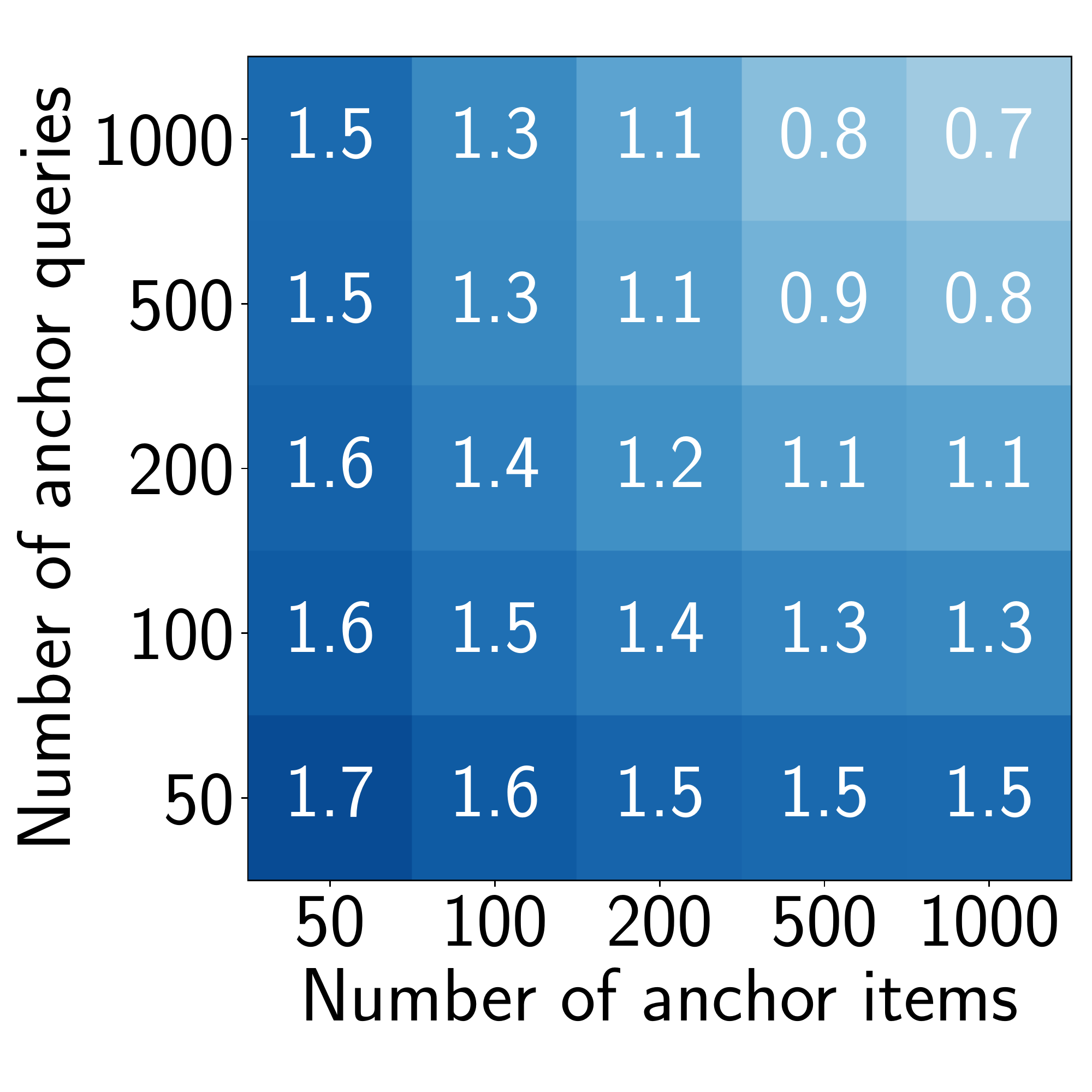}
         \caption{Top-10-Recall@Cost=500}
         \label{fig:arch_change_yugioh_prec_at_k_heatmap_ecrossenc_oracle_cur}
     \end{subfigure}
    \hfill
     \begin{subfigure}[b]{0.23\textwidth}
         \centering
         \includegraphics[width=\textwidth]{figure/rq_7_2_arch_change_yugioh_approx_error_heatmap_top_10_at_500_cur_oracle_ece.pdf}
         \caption{Matrix Approx. Error}
         \label{fig:arch_change_yugioh_approx_error_heatmap_ecrossenc_oracle_cur}
     \end{subfigure}
     \hfill
    \caption{
     Performance of \propMethod on non-anchor/test queries on domain =  \yugioh 
     using $U~=~C^\dagger M R^\dagger$ for \eCrossenc. The total number of queries including both anchor and non-anchor (test)
    queries is 3374 and the total number of items is 10031.}
    \label{fig:arch_change_yugioh_approx_error_heatmap_oracle_cur}
\end{figure}

Figure~\ref{fig:arch_change_yugioh_prec_at_k_heatmap} in~\S\ref{subsec:cur_analysis} shows
Top-10-Recall@500 on domain=\yugioh  for \clsCrossenc and \eCrossenc respectively
on different combinations of number of anchor 
queries~($\nAnchorQueries$) and anchor items~($\nAnchorItems$). 
Note that the performance of \propMethod drops significantly when $\nAnchorQueries = \nAnchorItems$.
Recall that the indexing step~(\S\ref{subsec:indexing}) requires computing
pseudo-inverse of matrix $A~=~M[\anchorQueries, \anchorItems]$ 
containing scores between anchor queries ($\anchorQueries$) and anchor items ($\anchorItems$).
Performance drops significantly when $A$ is a square matrix i.e. $\nAnchorQueries = \nAnchorItems$ 
as the matrix
tends to be ill-conditioned, with several very small eigenvalues 
that are ‘blown up’ in $A^{\dagger}$, the pseudo-inverse of $A$. This, in-turn, 
leads to a significant approximation error~\cite{archan2022sublinear}.
Choosing different number of anchor queries and anchor items yields a rectangular
matrix $A$ whose eigenvalues are unlikely to be small, thus resulting in much better 
approximation of the matrix $M$.

\paragraph{Oracle CUR Decomposition}
An alternate way of computing the matrix $U$ in CUR decomposition 
of a matrix $M$ for a given
subset of rows ($R$) and columns ($C$) is to set 
$U=C^{\dagger} M R^{\dagger} $.
This can provide a much stable approximation of 
the matrix $M$ even when $\nAnchorQueries=\nAnchorItems$~\cite{mahoney2009cur}.
However, it requires computing \emph{all} values of $M$ before
computing its low-rank approximation. In our case, we are trying to approximate
a matrix $M$ which also contains scores between test-queries and all items in order to
avoid scoring all items using the CE model at test-time, 
thus we can not use $U = C^{\dagger} M R^{\dagger}$.
Figure~\ref{fig:arch_change_yugioh_approx_error_heatmap_oracle_cur}
shows results for an oracle experiment where we use $U = C^{\dagger} M R^{\dagger}$,
and as expected it provides significant improvement when $\nAnchorQueries = \nAnchorItems$
and minor improvement otherwise over using $U = M[\anchorQueries, \anchorItems]^{\dagger}$.

\subsection{$k$-NN experiment results for all domains}
\label{subsec:appendix_exhaustive_exp_res}
\begin{figure*}[]
     \centering
     \begin{subfigure}[b]{0.95\textwidth}
        \centering
        \includegraphics[width=\textwidth]{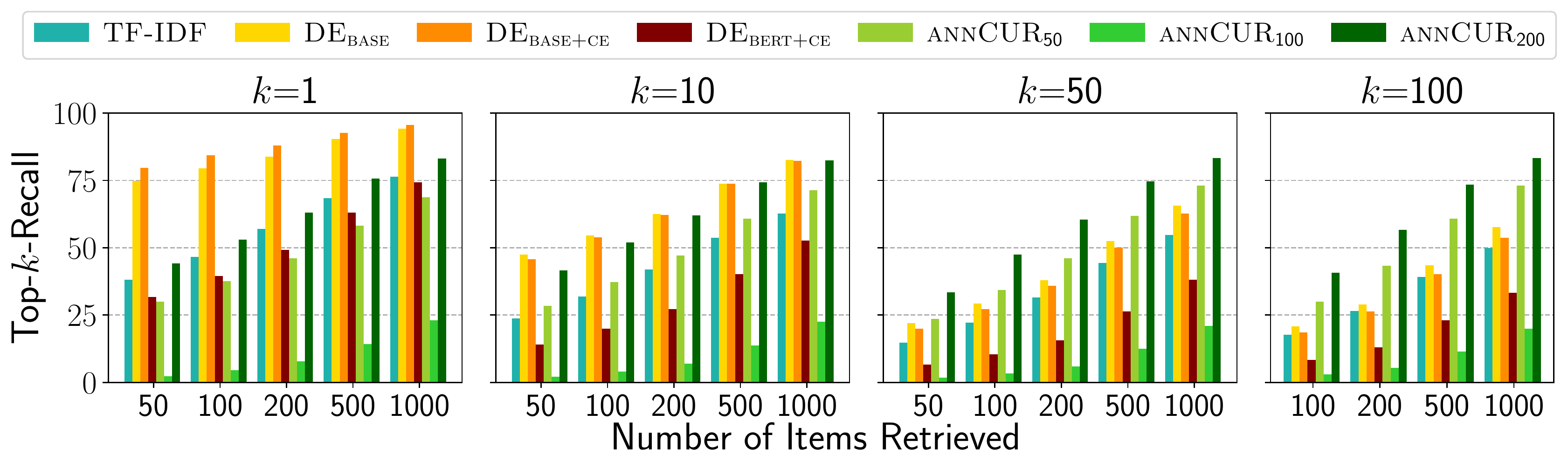}
        \caption{Top-$k$-Recall@$\nRetrievedItems$ for \propMethod and baselines when all methods retrieve and rerank 
        the same number of items $(\nRetrievedItems)$. The subscript $\nAnchorItems$ in $\propMethod_{\nAnchorItems}$ refers to
        the number of anchor items used for embedding the test query.}
        \label{fig:appendix_recall_at_same_k_retrieved_yugioh_100}
    \end{subfigure}
     \hfill
     \begin{subfigure}[b]{0.95\textwidth}
    \centering
    \includegraphics[width=\textwidth]{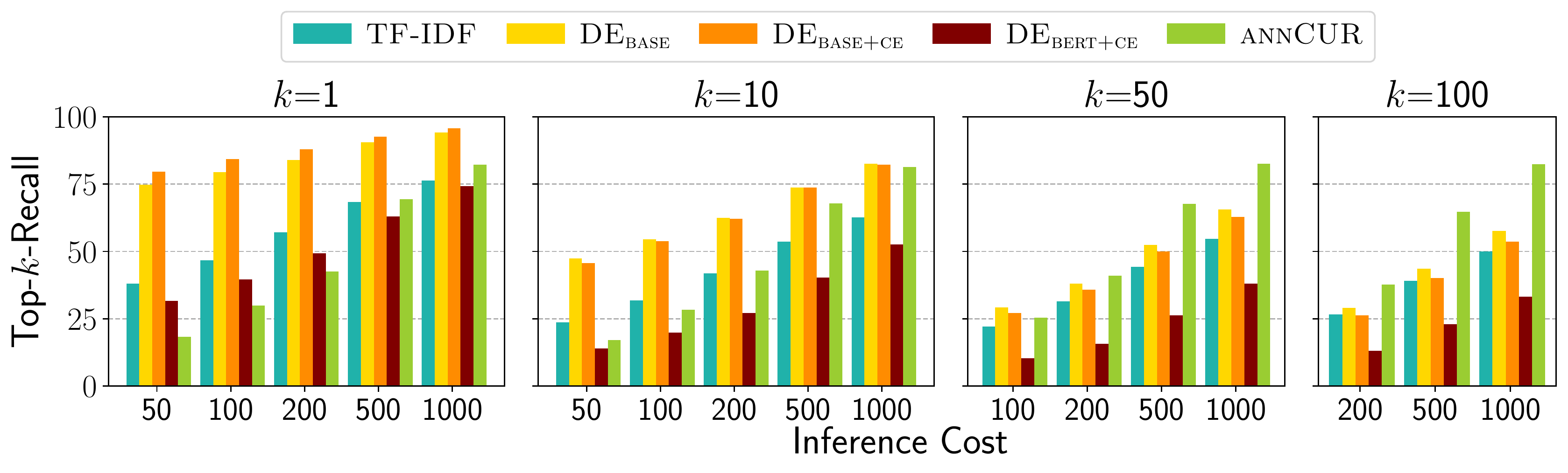}
    \caption{Top-$k$-Recall for \propMethod and baselines  when all methods operate under a fixed test-time cost budget. Recall 
    that cost is the number of CE calls made during inference
    for re-ranking retrieved items and, in case of \propMethod, it also includes CE calls to embed the test query by comparing with anchor items.
    }
    \label{fig:appendix_recall_at_same_cost_yugioh_100}
\end{subfigure}
    \caption{Top-$k$-Recall results for domain=\yugioh and $\queryTrainSize=100$}
    \label{fig:appendix_recall_yugioh_100}
\end{figure*}

\begin{figure*}[]
     \centering
     \begin{subfigure}[b]{0.95\textwidth}
        \centering
        \includegraphics[width=\textwidth]{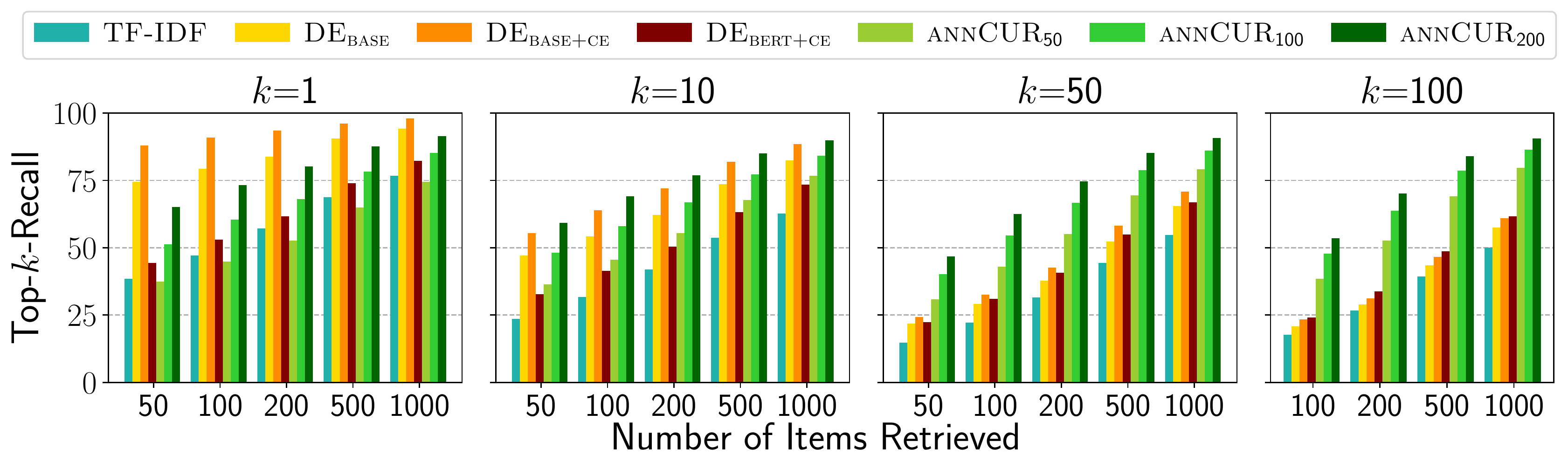}
        \caption{Top-$k$-Recall@$\nRetrievedItems$ for \propMethod and baselines when all methods retrieve and rerank 
        the same number of items $(\nRetrievedItems)$. The subscript $\nAnchorItems$ in $\propMethod_{\nAnchorItems}$ refers to
        the number of anchor items used for embedding the test query.}
        \label{fig:appendix_recall_at_same_k_retrieved_yugioh_500}
    \end{subfigure}
     \hfill
     \begin{subfigure}[b]{0.95\textwidth}
    \centering
    \includegraphics[width=\textwidth]{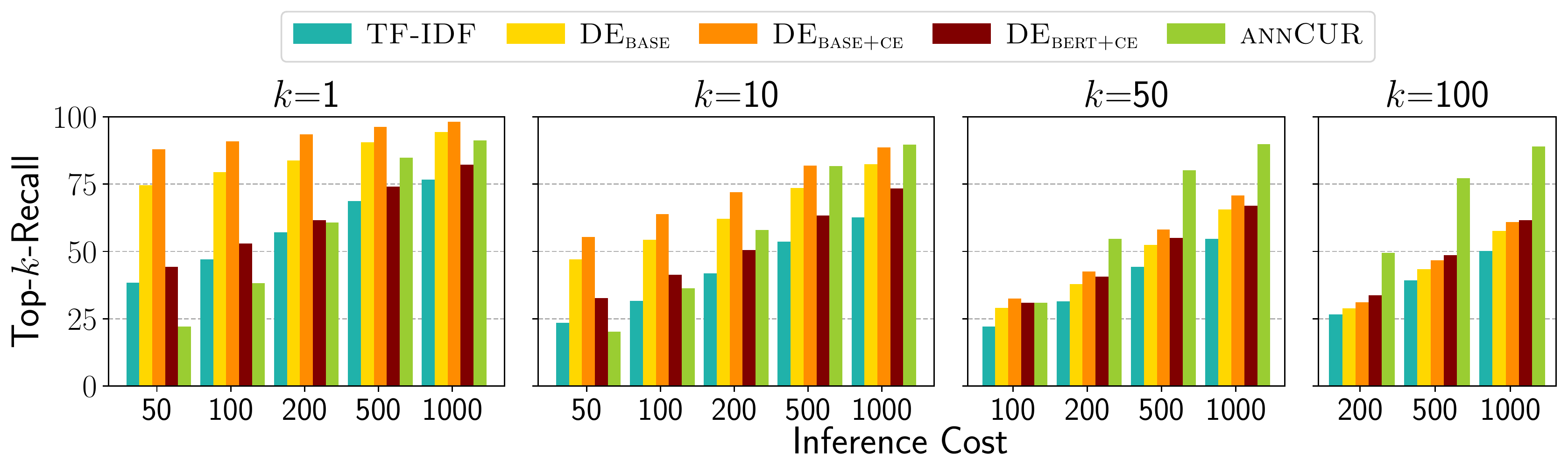}
    \caption{Top-$k$-Recall for \propMethod and baselines  when all methods operate under a fixed test-time cost budget. Recall 
    that cost is the number of CE calls made during inference
    for re-ranking retrieved items and, in case of \propMethod, it also includes CE calls to embed the test query by comparing with anchor items.
    }
    \label{fig:appendix_recall_at_same_cost_yugioh_500}
\end{subfigure}
    \caption{Top-$k$-Recall results for domain=\yugioh and $\queryTrainSize=500$}
\end{figure*}

\begin{figure*}[]
     \centering
     \begin{subfigure}[b]{0.95\textwidth}
        \centering
        \includegraphics[width=\textwidth]{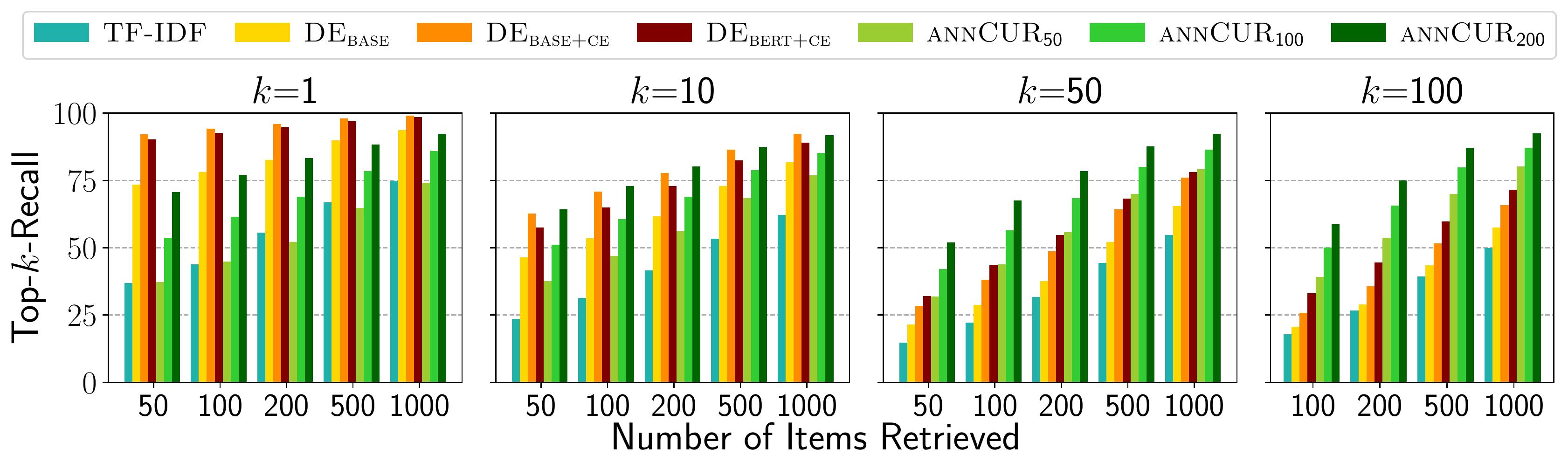}
        \caption{Top-$k$-Recall@$\nRetrievedItems$ for \propMethod and baselines when all methods retrieve and rerank 
        the same number of items $(\nRetrievedItems)$. The subscript $\nAnchorItems$ in $\propMethod_{\nAnchorItems}$ refers to
        the number of anchor items used for embedding the test query.}
        \label{fig:appendix_recall_at_same_k_retrieved_yugioh_2000}
    \end{subfigure}
     \hfill
     \begin{subfigure}[b]{0.95\textwidth}
    \centering
    \includegraphics[width=\textwidth]{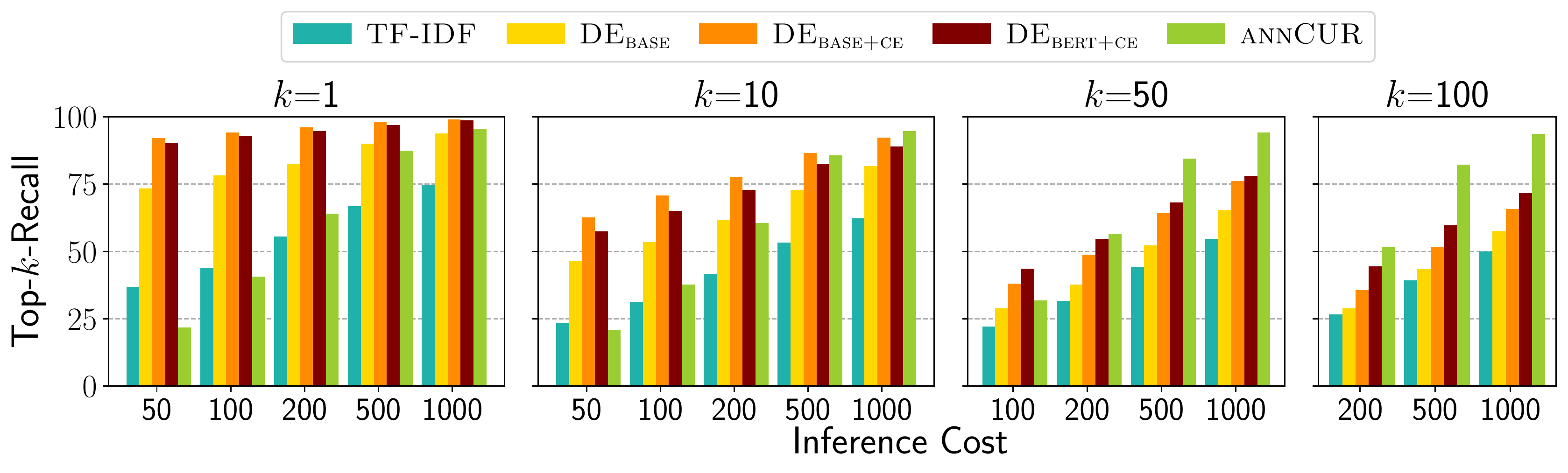}
    \caption{Top-$k$-Recall for \propMethod and baselines  when all methods operate under a fixed test-time cost budget. Recall 
    that cost is the number of CE calls made during inference
    for re-ranking retrieved items and, in case of \propMethod, it also includes CE calls to embed the test query by comparing with anchor items.
    }
    \label{fig:appendix_recall_at_same_cost_yugioh_2000}
\end{subfigure}
    \caption{Top-$k$-Recall results for domain=\yugioh and $\queryTrainSize=2000$}
\end{figure*}

\begin{figure*}[]
     \centering
     \begin{subfigure}[b]{0.95\textwidth}
        \centering
        \includegraphics[width=\textwidth]{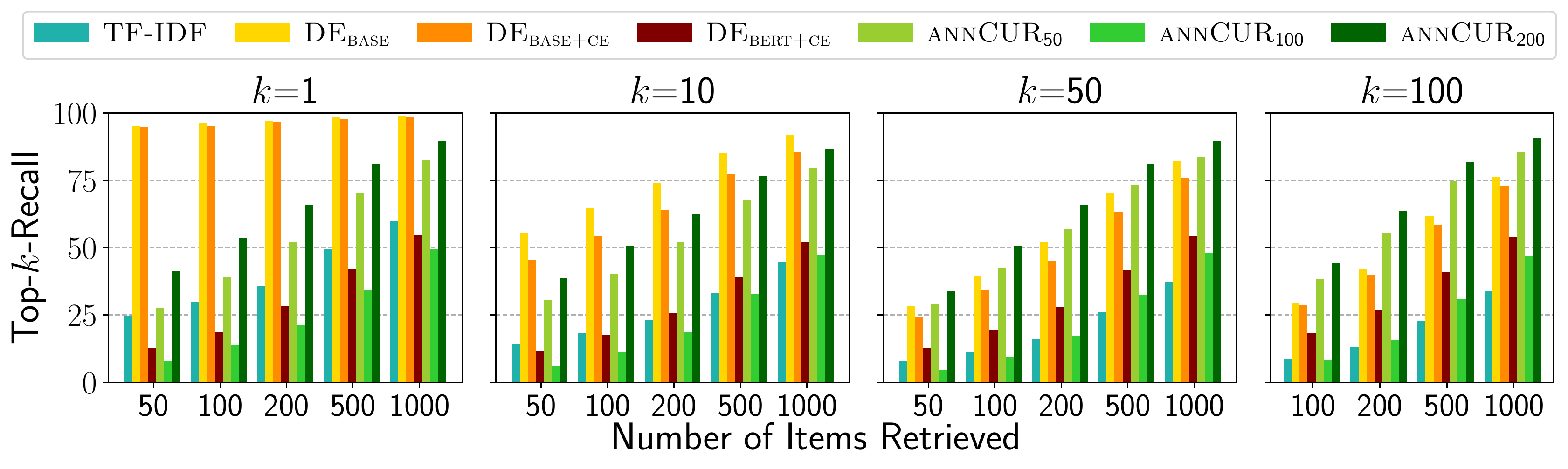}
        \caption{Top-$k$-Recall@$\nRetrievedItems$ for \propMethod and baselines when all methods retrieve and rerank 
        the same number of items $(\nRetrievedItems)$. The subscript $\nAnchorItems$ in $\propMethod_{\nAnchorItems}$ refers to
        the number of anchor items used for embedding the test query.}
        \label{fig:appendix_recall_at_same_k_retrieved_pro_wrestling_100}
    \end{subfigure}
     \hfill
     \begin{subfigure}[b]{0.95\textwidth}
    \centering
    \includegraphics[width=\textwidth]{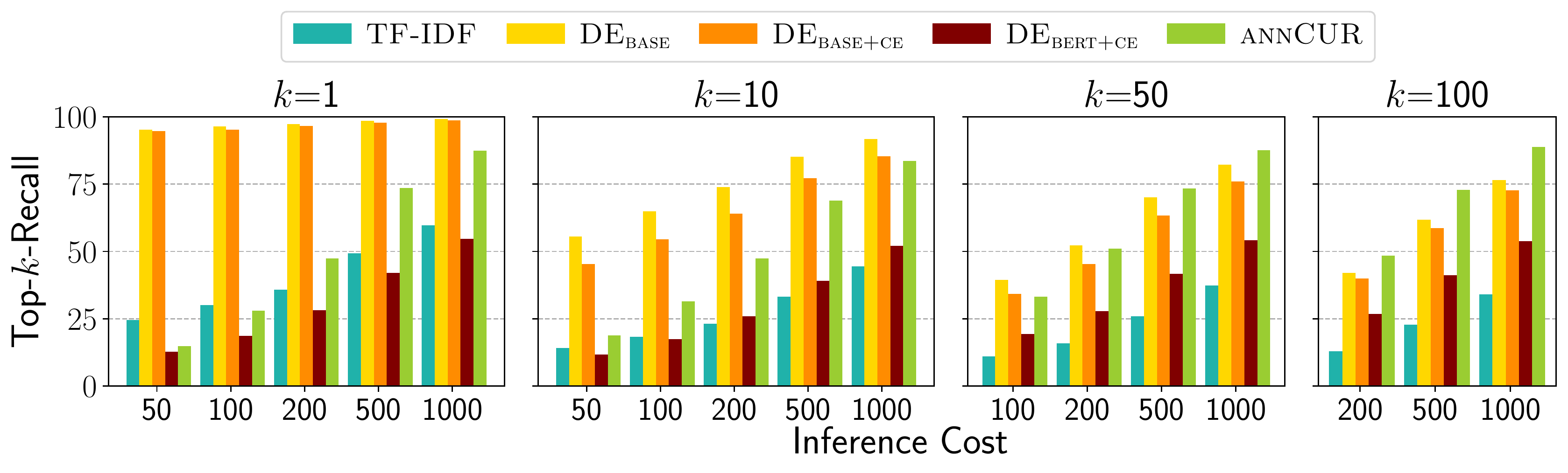}
    \caption{Top-$k$-Recall for \propMethod and baselines  when all methods operate under a fixed test-time cost budget. Recall 
    that cost is the number of CE calls made during inference
    for re-ranking retrieved items and, in case of \propMethod, it also includes CE calls to embed the test query by comparing with anchor items.
    }
    \label{fig:appendix_recall_at_same_cost_pro_wrestling_100}
\end{subfigure}
    \caption{Top-$k$-Recall results for domain=\proWrestling and $\queryTrainSize=100$}
\end{figure*}

\begin{figure*}[]
     \centering
     \begin{subfigure}[b]{0.95\textwidth}
        \centering
        \includegraphics[width=\textwidth]{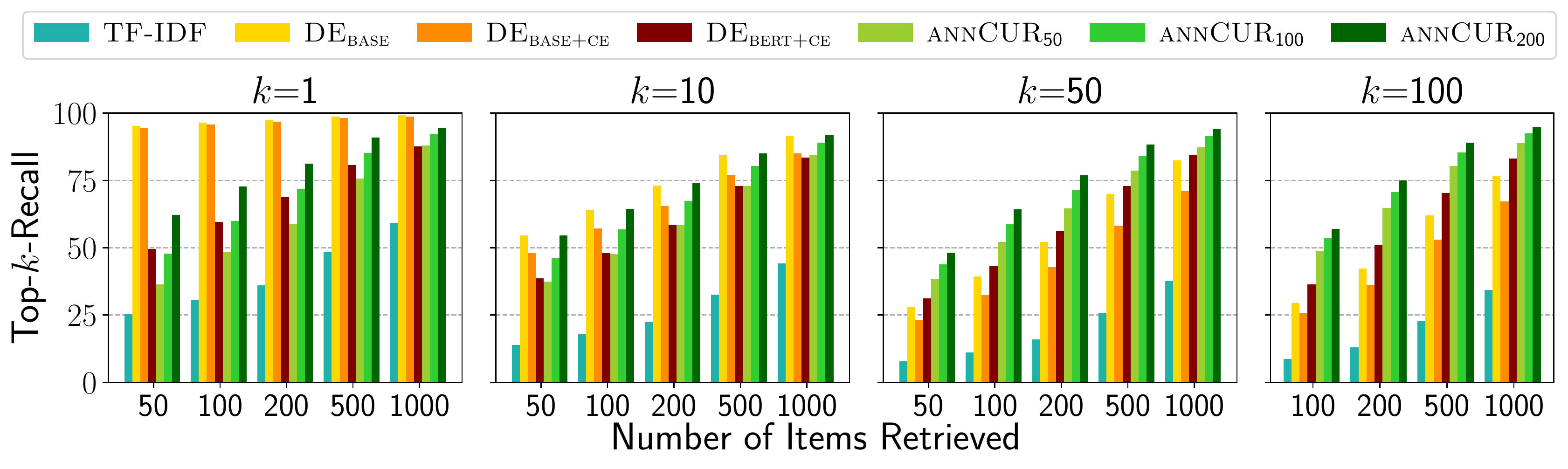}
        \caption{Top-$k$-Recall@$\nRetrievedItems$ for \propMethod and baselines when all methods retrieve and rerank 
        the same number of items $(\nRetrievedItems)$. The subscript $\nAnchorItems$ in $\propMethod_{\nAnchorItems}$ refers to
        the number of anchor items used for embedding the test query.}
        \label{fig:appendix_recall_at_same_k_retrieved_pro_wrestling_500}
    \end{subfigure}
     \hfill
     \begin{subfigure}[b]{0.95\textwidth}
    \centering
    \includegraphics[width=\textwidth]{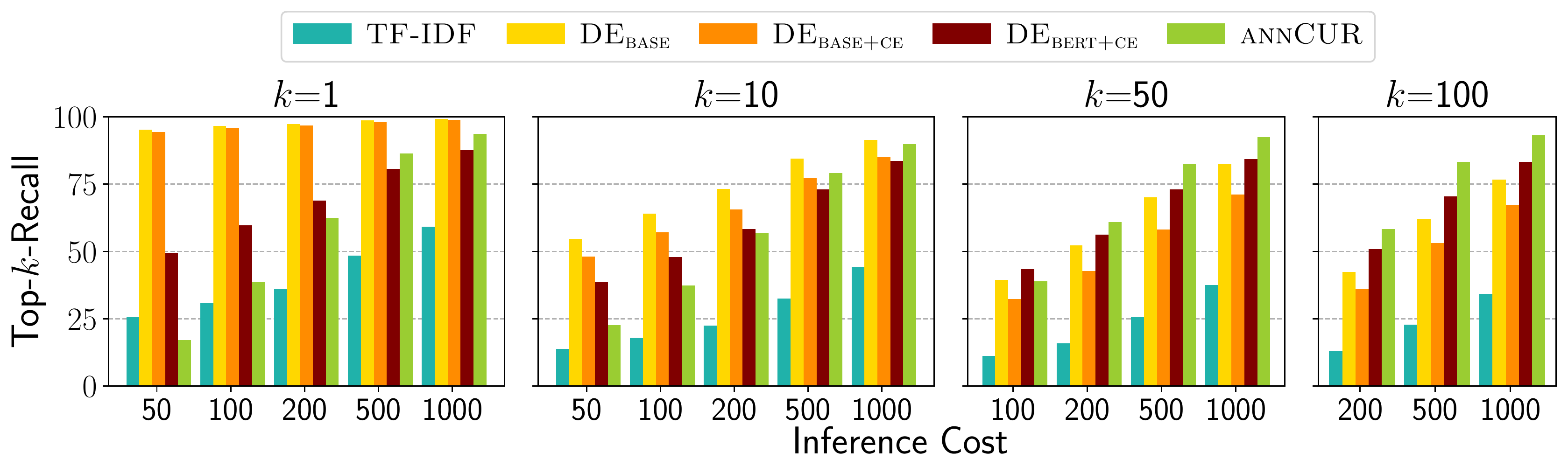}
    \caption{Top-$k$-Recall for \propMethod and baselines  when all methods operate under a fixed test-time cost budget. Recall 
    that cost is the number of CE calls made during inference
    for re-ranking retrieved items and, in case of \propMethod, it also includes CE calls to embed the test query by comparing with anchor items.
    }
    \label{fig:appendix_recall_at_same_cost_pro_wrestling_500}
\end{subfigure}
    \caption{Top-$k$-Recall results for domain=\proWrestling and $\queryTrainSize=500$}
\end{figure*}

\begin{figure*}[]
     \centering
     \begin{subfigure}[b]{0.95\textwidth}
        \centering
        \includegraphics[width=\textwidth]{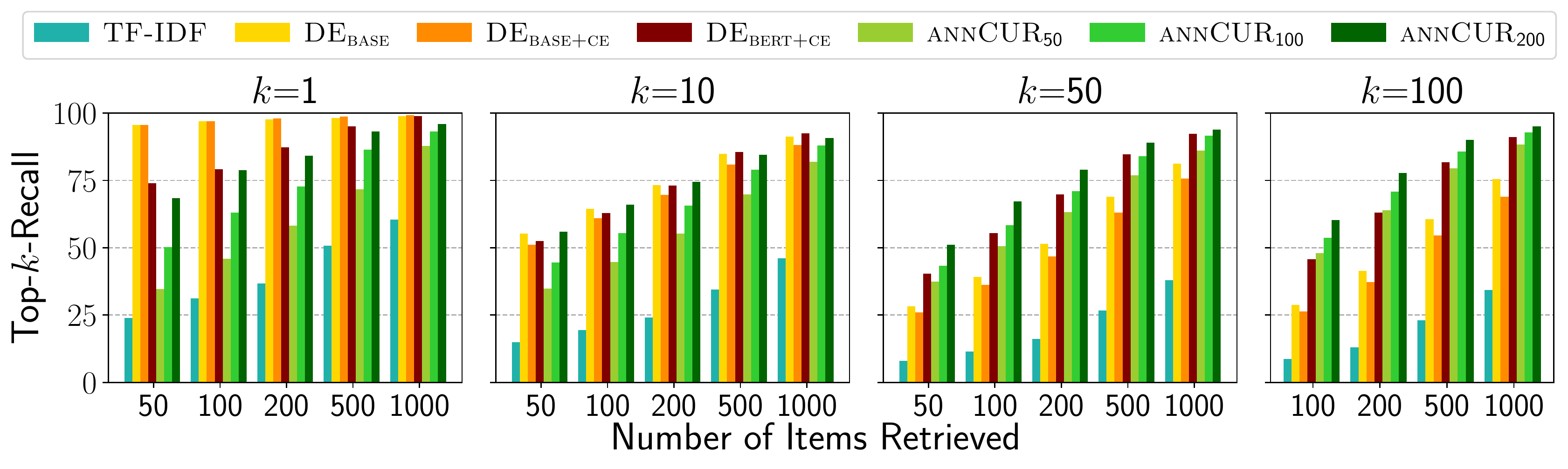}
        \caption{Top-$k$-Recall@$\nRetrievedItems$ for \propMethod and baselines when all methods retrieve and rerank 
        the same number of items $(\nRetrievedItems)$. The subscript $\nAnchorItems$ in $\propMethod_{\nAnchorItems}$ refers to
        the number of anchor items used for embedding the test query.}
        \label{fig:appendix_recall_at_same_k_retrieved_pro_wrestling_1000}
    \end{subfigure}
     \hfill
     \begin{subfigure}[b]{0.95\textwidth}
    \centering
    \includegraphics[width=\textwidth]{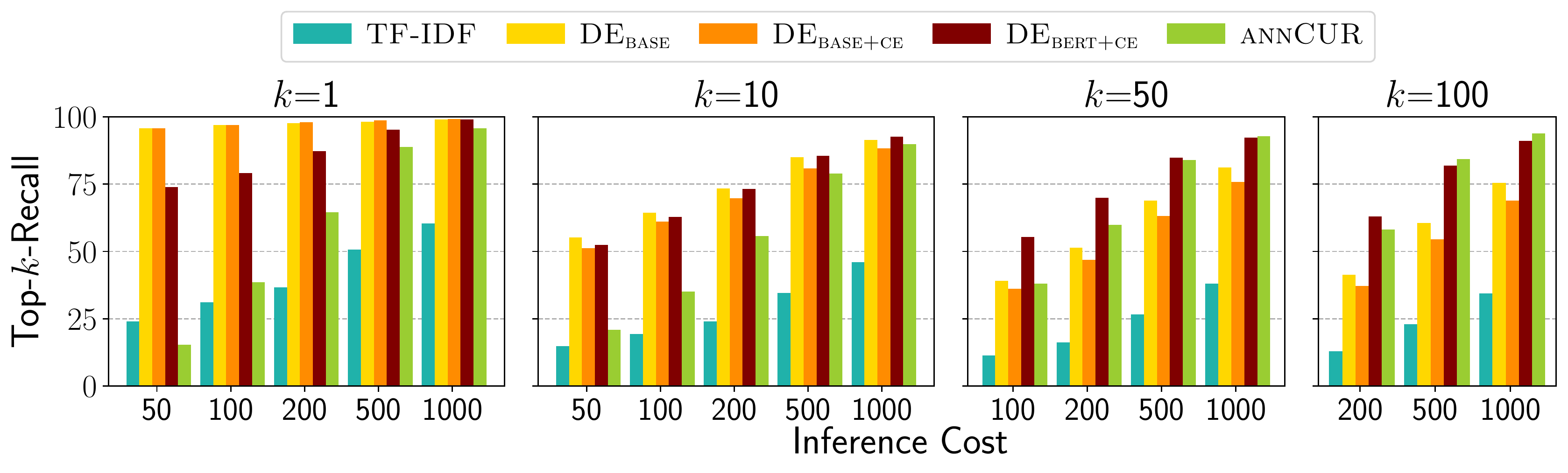}
    \caption{Top-$k$-Recall for \propMethod and baselines  when all methods operate under a fixed test-time cost budget. Recall 
    that cost is the number of CE calls made during inference
    for re-ranking retrieved items and, in case of \propMethod, it also includes CE calls to embed the test query by comparing with anchor items.
    }
    \label{fig:appendix_recall_at_same_cost_pro_wrestling_1000}
\end{subfigure}
    \caption{Top-$k$-Recall results for domain=\proWrestling and $\queryTrainSize=1000$}
\end{figure*}

\begin{figure*}[]
     \centering
     \begin{subfigure}[b]{0.95\textwidth}
        \centering
        \includegraphics[width=\textwidth]{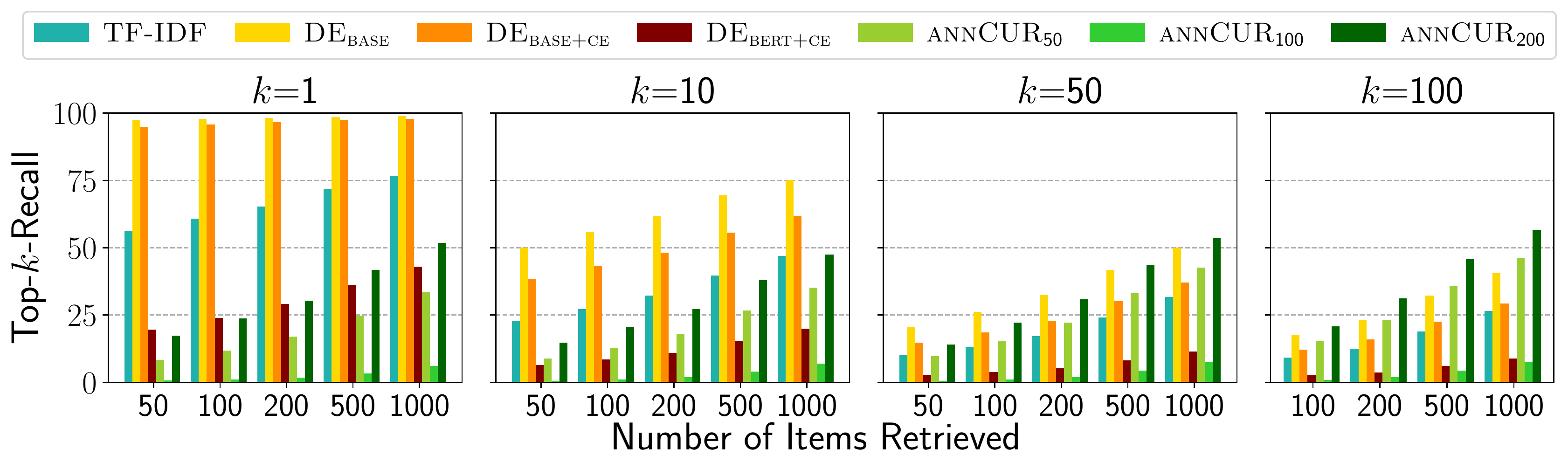}
        \caption{Top-$k$-Recall@$\nRetrievedItems$ for \propMethod and baselines when all methods retrieve and rerank 
        the same number of items $(\nRetrievedItems)$. The subscript $\nAnchorItems$ in $\propMethod_{\nAnchorItems}$ refers to
        the number of anchor items used for embedding the test query.}
        \label{fig:appendix_recall_at_same_k_retrieved_doctor_who_100}
    \end{subfigure}
     \hfill
     \begin{subfigure}[b]{0.95\textwidth}
    \centering
    \includegraphics[width=\textwidth]{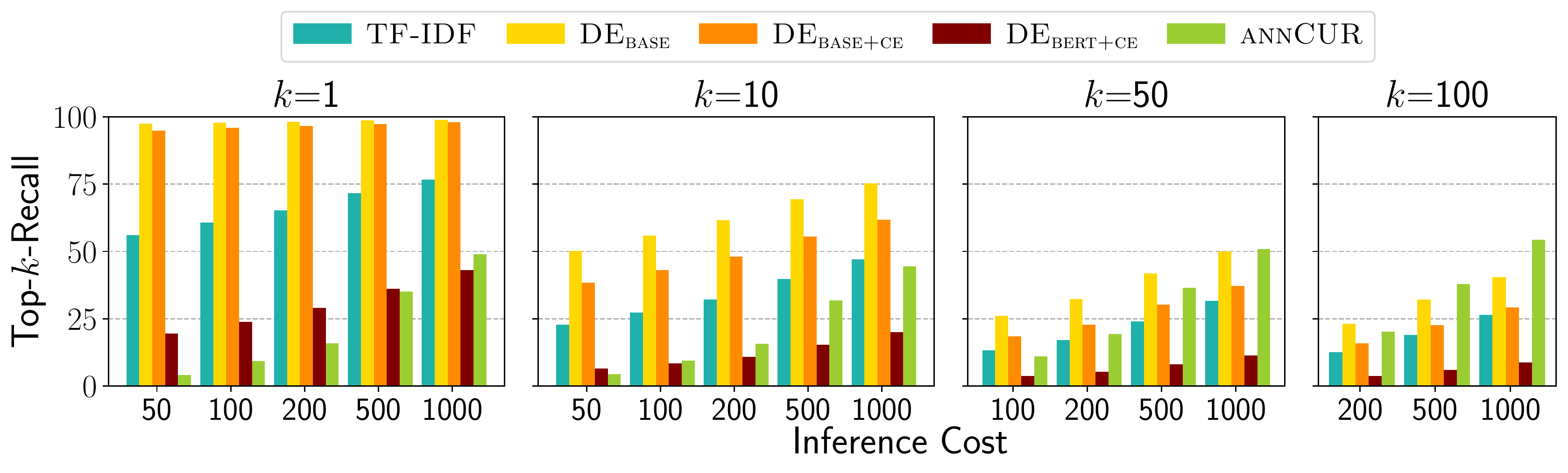}
    \caption{Top-$k$-Recall for \propMethod and baselines  when all methods operate under a fixed test-time cost budget. Recall 
    that cost is the number of CE calls made during inference
    for re-ranking retrieved items and, in case of \propMethod, it also includes CE calls to embed the test query by comparing with anchor items.
    }
    \label{fig:appendix_recall_at_same_cost_doctor_who_100}
\end{subfigure}
    \caption{Top-$k$-Recall results for domain=\doctorWho and $\queryTrainSize=100$}
\end{figure*}

\begin{figure*}[]
     \centering
     \begin{subfigure}[b]{0.95\textwidth}
        \centering
        \includegraphics[width=\textwidth]{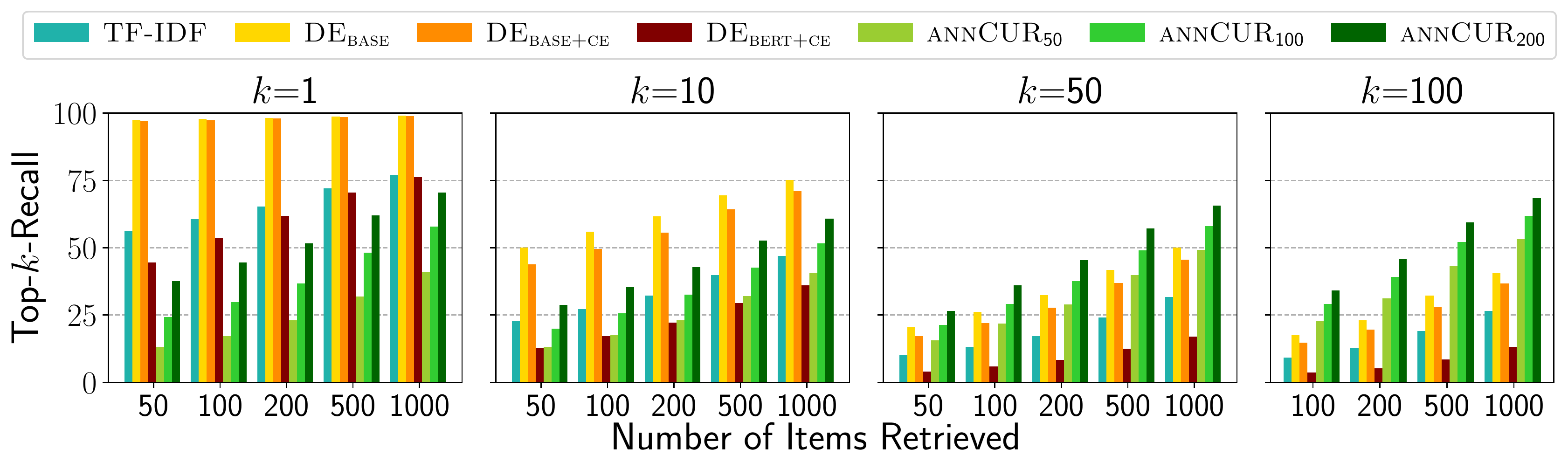}
        \caption{Top-$k$-Recall@$\nRetrievedItems$ for \propMethod and baselines when all methods retrieve and rerank 
        the same number of items $(\nRetrievedItems)$. The subscript $\nAnchorItems$ in $\propMethod_{\nAnchorItems}$ refers to
        the number of anchor items used for embedding the test query.}
        \label{fig:appendix_recall_at_same_k_retrieved_doctor_who_500}
    \end{subfigure}
     \hfill
     \begin{subfigure}[b]{0.95\textwidth}
    \centering
    \includegraphics[width=\textwidth]{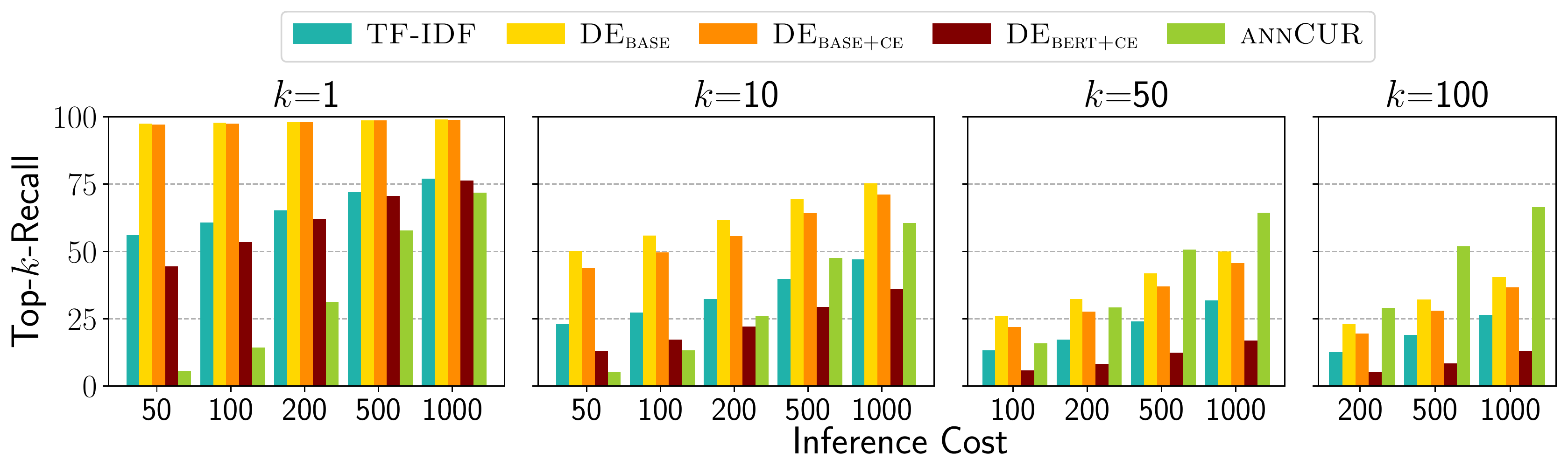}
    \caption{Top-$k$-Recall for \propMethod and baselines  when all methods operate under a fixed test-time cost budget. Recall 
    that cost is the number of CE calls made during inference
    for re-ranking retrieved items and, in case of \propMethod, it also includes CE calls to embed the test query by comparing with anchor items.
    }
    \label{fig:appendix_recall_at_same_cost_doctor_who_500}
\end{subfigure}
    \caption{Top-$k$-Recall results for domain=\doctorWho and $\queryTrainSize=500$}
\end{figure*}

\begin{figure*}[]
     \centering
     \begin{subfigure}[b]{0.95\textwidth}
        \centering
        \includegraphics[width=\textwidth]{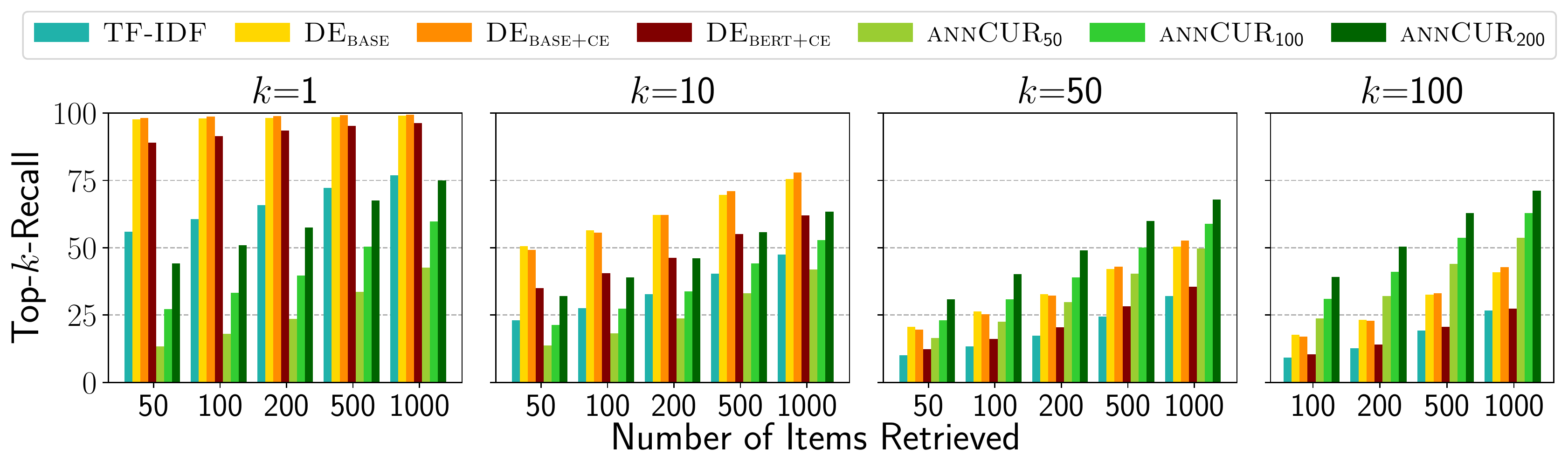}
        \caption{Top-$k$-Recall@$\nRetrievedItems$ for \propMethod and baselines when all methods retrieve and rerank 
        the same number of items $(\nRetrievedItems)$. The subscript $\nAnchorItems$ in $\propMethod_{\nAnchorItems}$ refers to
        the number of anchor items used for embedding the test query.}
        \label{fig:appendix_recall_at_same_k_retrieved_doctor_who_2000}
    \end{subfigure}
     \hfill
     \begin{subfigure}[b]{0.95\textwidth}
    \centering
    \includegraphics[width=\textwidth]{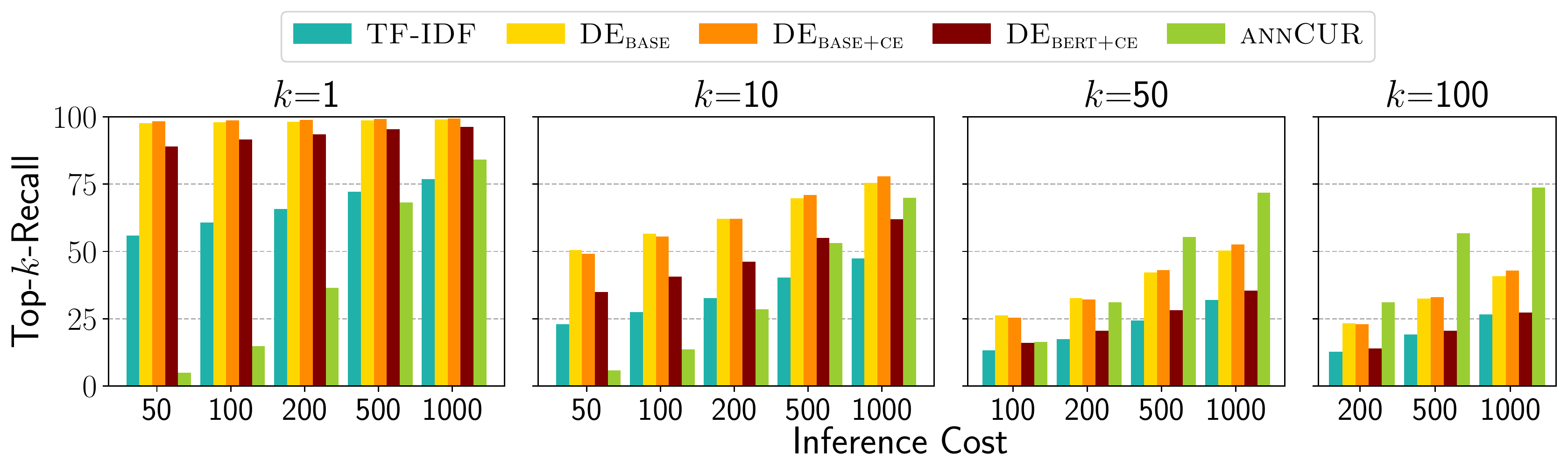}
    \caption{Top-$k$-Recall for \propMethod and baselines  when all methods operate under a fixed test-time cost budget. Recall 
    that cost is the number of CE calls made during inference
    for re-ranking retrieved items and, in case of \propMethod, it also includes CE calls to embed the test query by comparing with anchor items.
    }
    \label{fig:appendix_recall_at_same_cost_doctor_who_2000}
\end{subfigure}
    \caption{Top-$k$-Recall results for domain=\doctorWho and $\queryTrainSize=2000$}
\end{figure*}

\begin{figure*}[]
     \centering
     \begin{subfigure}[b]{0.95\textwidth}
        \centering
        \includegraphics[width=\textwidth]{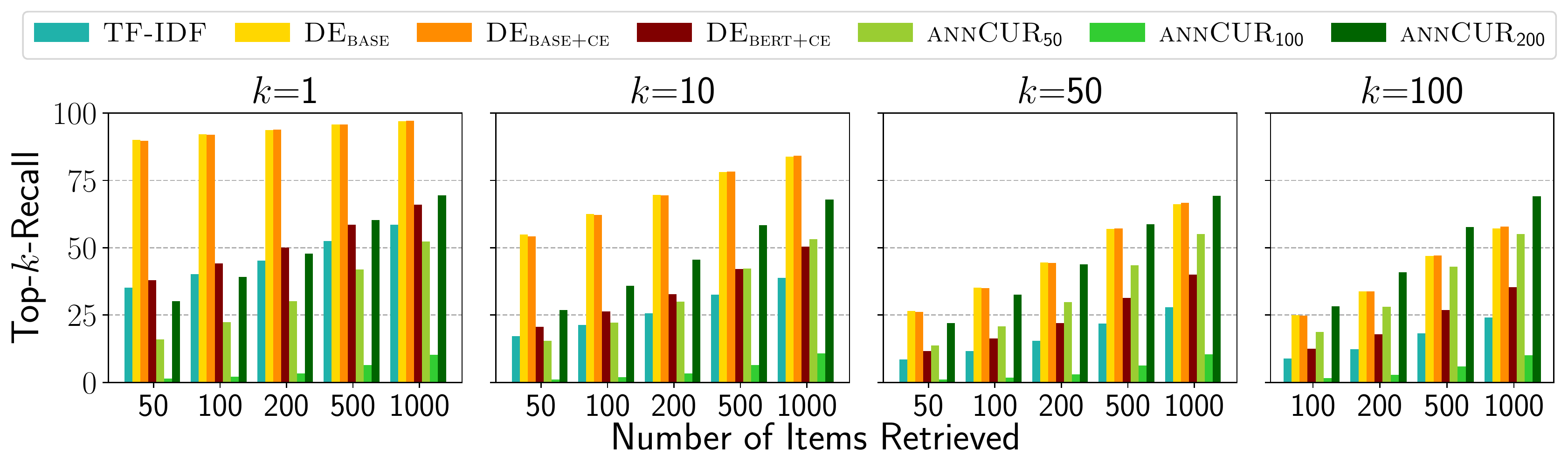}
        \caption{Top-$k$-Recall@$\nRetrievedItems$ for \propMethod and baselines when all methods retrieve and rerank 
        the same number of items $(\nRetrievedItems)$. The subscript $\nAnchorItems$ in $\propMethod_{\nAnchorItems}$ refers to
        the number of anchor items used for embedding the test query.}
        \label{fig:appendix_recall_at_same_k_retrieved_star_trek_100}
    \end{subfigure}
     \hfill
     \begin{subfigure}[b]{0.95\textwidth}
    \centering
    \includegraphics[width=\textwidth]{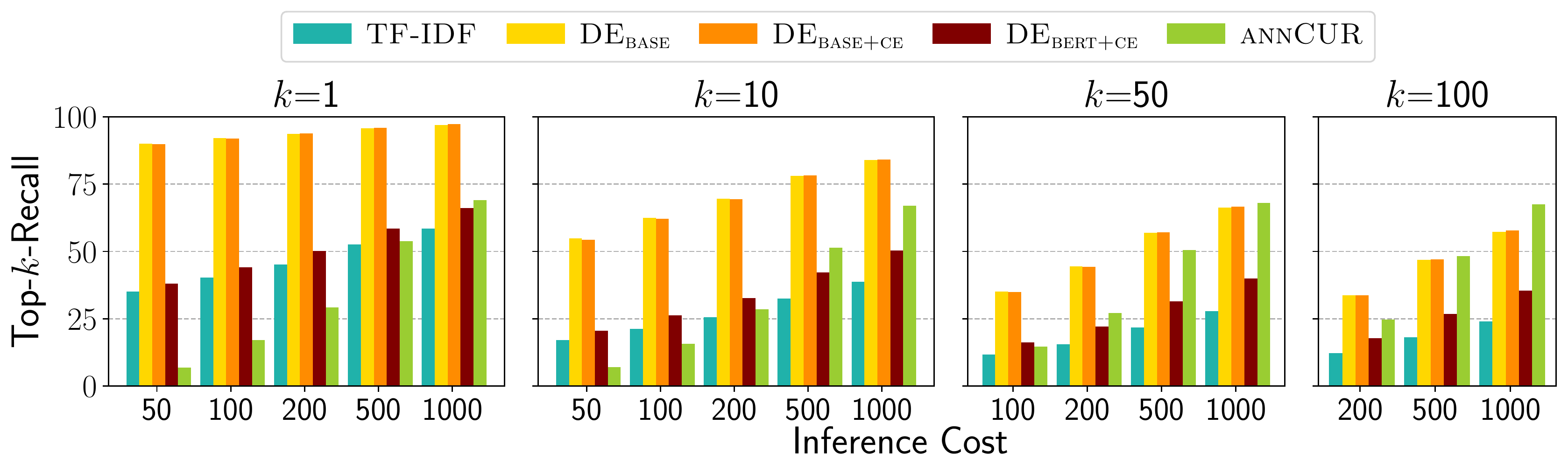}
    \caption{Top-$k$-Recall for \propMethod and baselines  when all methods operate under a fixed test-time cost budget. Recall 
    that cost is the number of CE calls made during inference
    for re-ranking retrieved items and, in case of \propMethod, it also includes CE calls to embed the test query by comparing with anchor items.
    }
    \label{fig:appendix_recall_at_same_cost_star_trek_100}
\end{subfigure}
    \caption{Top-$k$-Recall results for domain=\starTrek and $\queryTrainSize=100$}
\end{figure*}

\begin{figure*}[]
     \centering
     \begin{subfigure}[b]{0.95\textwidth}
        \centering
        \includegraphics[width=\textwidth]{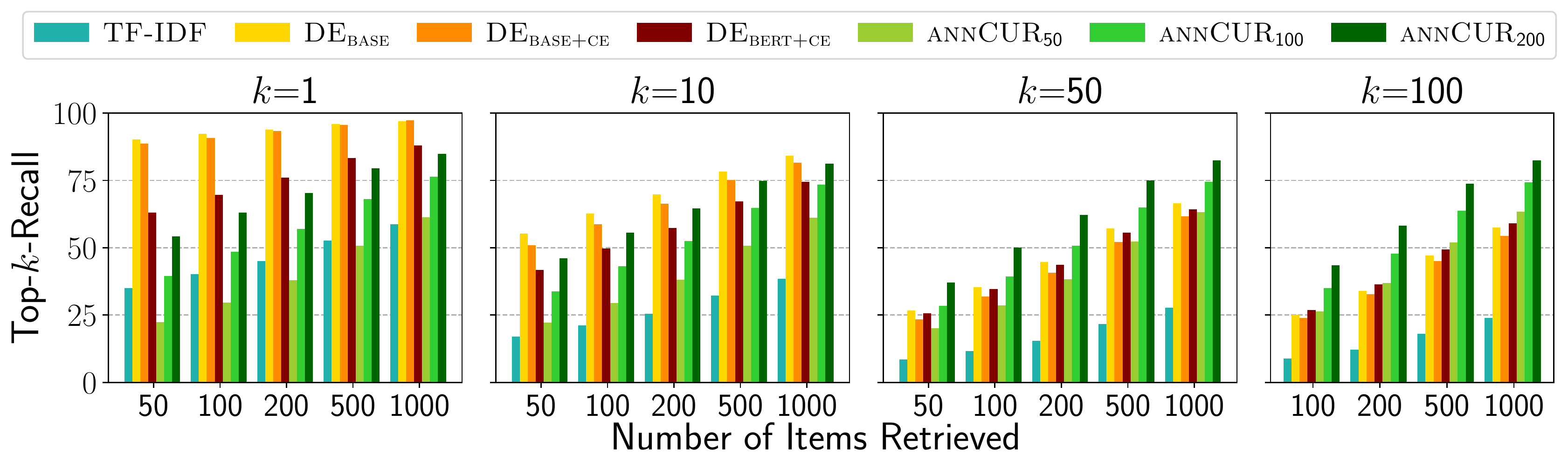}
        \caption{Top-$k$-Recall@$\nRetrievedItems$ for \propMethod and baselines when all methods retrieve and rerank 
        the same number of items $(\nRetrievedItems)$. The subscript $\nAnchorItems$ in $\propMethod_{\nAnchorItems}$ refers to
        the number of anchor items used for embedding the test query.}
        \label{fig:appendix_recall_at_same_k_retrieved_star_trek_500}
    \end{subfigure}
     \hfill
     \begin{subfigure}[b]{0.95\textwidth}
    \centering
    \includegraphics[width=\textwidth]{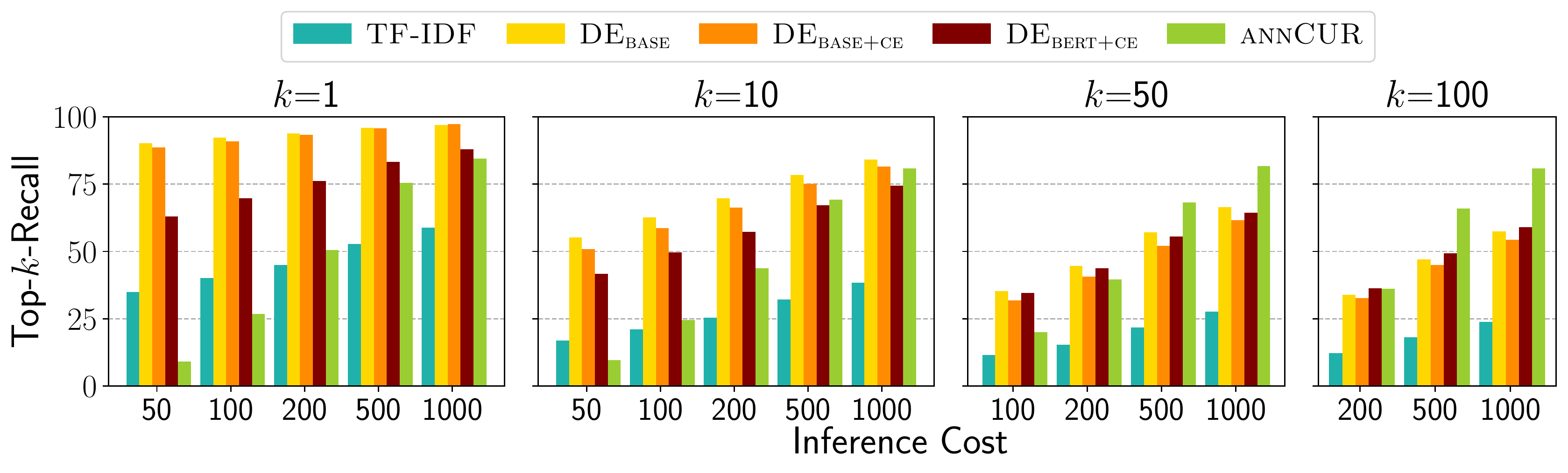}
    \caption{Top-$k$-Recall for \propMethod and baselines  when all methods operate under a fixed test-time cost budget. Recall 
    that cost is the number of CE calls made during inference
    for re-ranking retrieved items and, in case of \propMethod, it also includes CE calls to embed the test query by comparing with anchor items.
    }
    \label{fig:appendix_recall_at_same_cost_star_trek_500}
\end{subfigure}
    \caption{Top-$k$-Recall results for domain=\starTrek and $\queryTrainSize=500$}
\end{figure*}

\begin{figure*}[]
     \centering
     \begin{subfigure}[b]{0.95\textwidth}
        \centering
        \includegraphics[width=\textwidth]{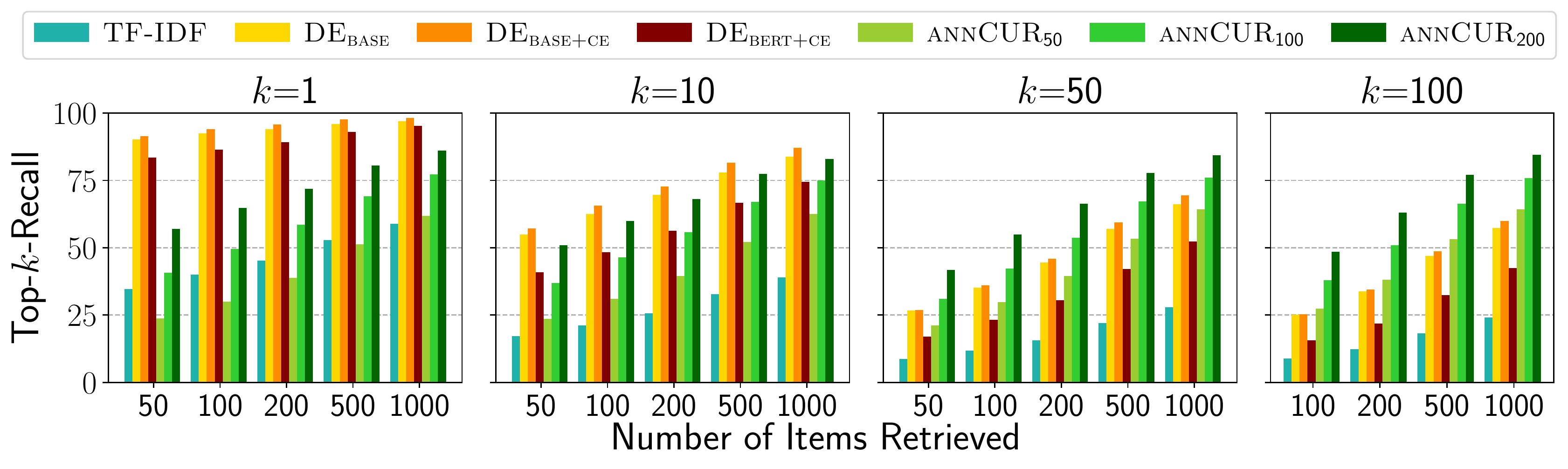}
        \caption{Top-$k$-Recall@$\nRetrievedItems$ for \propMethod and baselines when all methods retrieve and rerank 
        the same number of items $(\nRetrievedItems)$. The subscript $\nAnchorItems$ in $\propMethod_{\nAnchorItems}$ refers to
        the number of anchor items used for embedding the test query.}
        \label{fig:appendix_recall_at_same_k_retrieved_star_trek_2000}
    \end{subfigure}
     \hfill
     \begin{subfigure}[b]{0.95\textwidth}
    \centering
    \includegraphics[width=\textwidth]{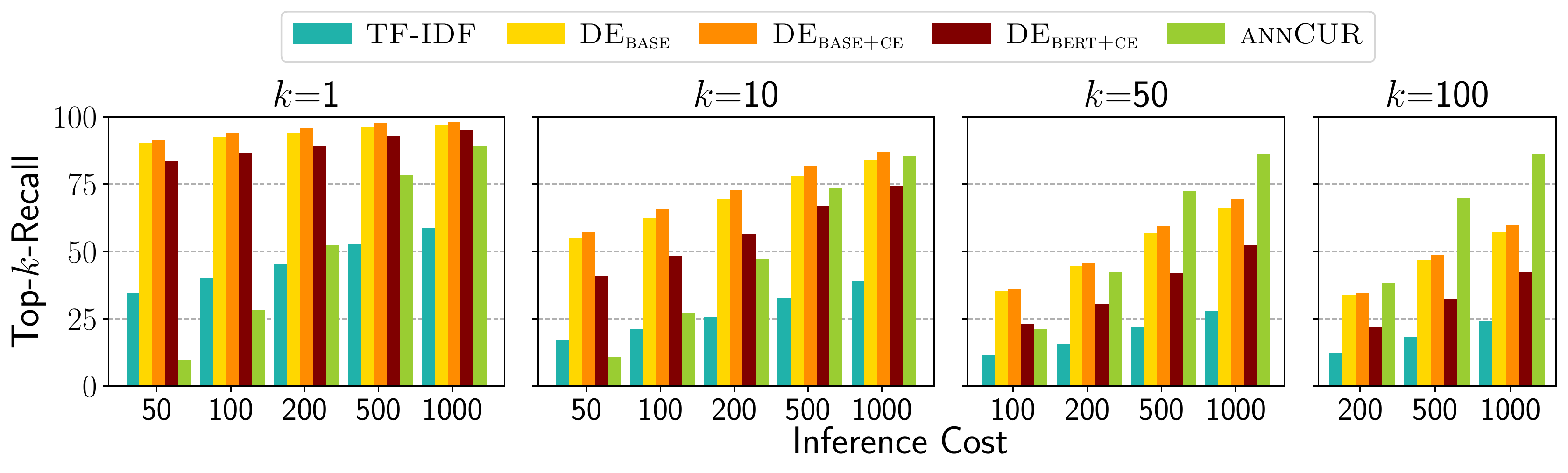}
    \caption{Top-$k$-Recall for \propMethod and baselines  when all methods operate under a fixed test-time cost budget. Recall 
    that cost is the number of CE calls made during inference
    for re-ranking retrieved items and, in case of \propMethod, it also includes CE calls to embed the test query by comparing with anchor items.
    }
    \label{fig:appendix_recall_at_same_cost_star_trek_2000}
\end{subfigure}
    \caption{Top-$k$-Recall results for domain=\starTrek and $\queryTrainSize=2000$}
\end{figure*}

\begin{figure*}[]
     \centering
     \begin{subfigure}[b]{0.95\textwidth}
        \centering
        \includegraphics[width=\textwidth]{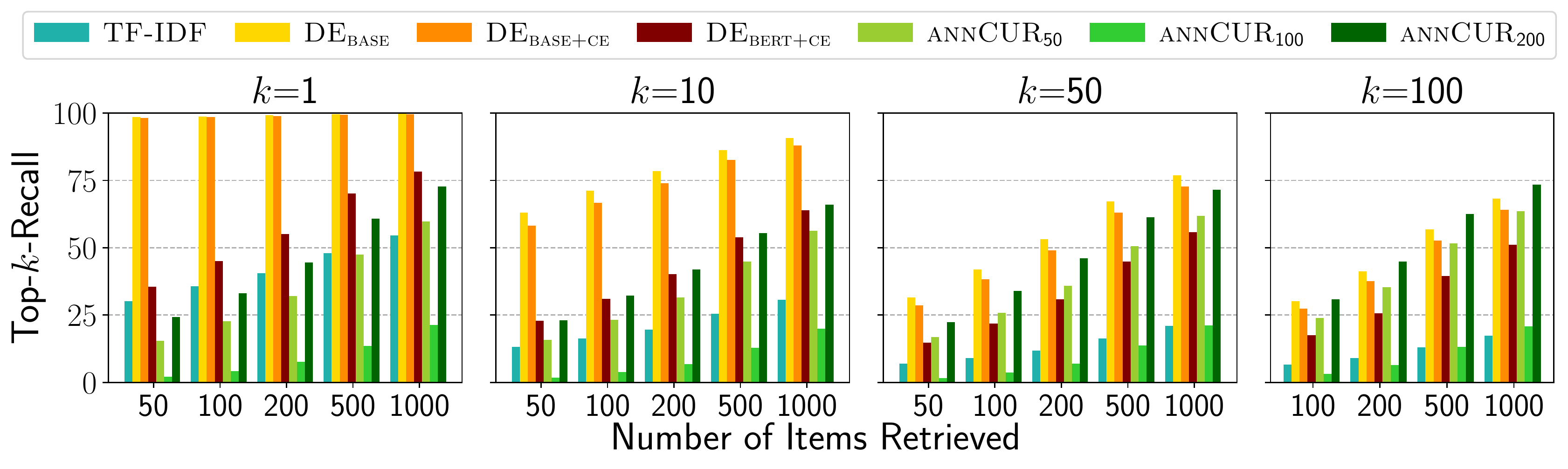}
        \caption{Top-$k$-Recall@$\nRetrievedItems$ for \propMethod and baselines when all methods retrieve and rerank 
        the same number of items $(\nRetrievedItems)$. The subscript $\nAnchorItems$ in $\propMethod_{\nAnchorItems}$ refers to
        the number of anchor items used for embedding the test query.}
        \label{fig:appendix_recall_at_same_k_retrieved_military_100}
    \end{subfigure}
     \hfill
     \begin{subfigure}[b]{0.95\textwidth}
    \centering
    \includegraphics[width=\textwidth]{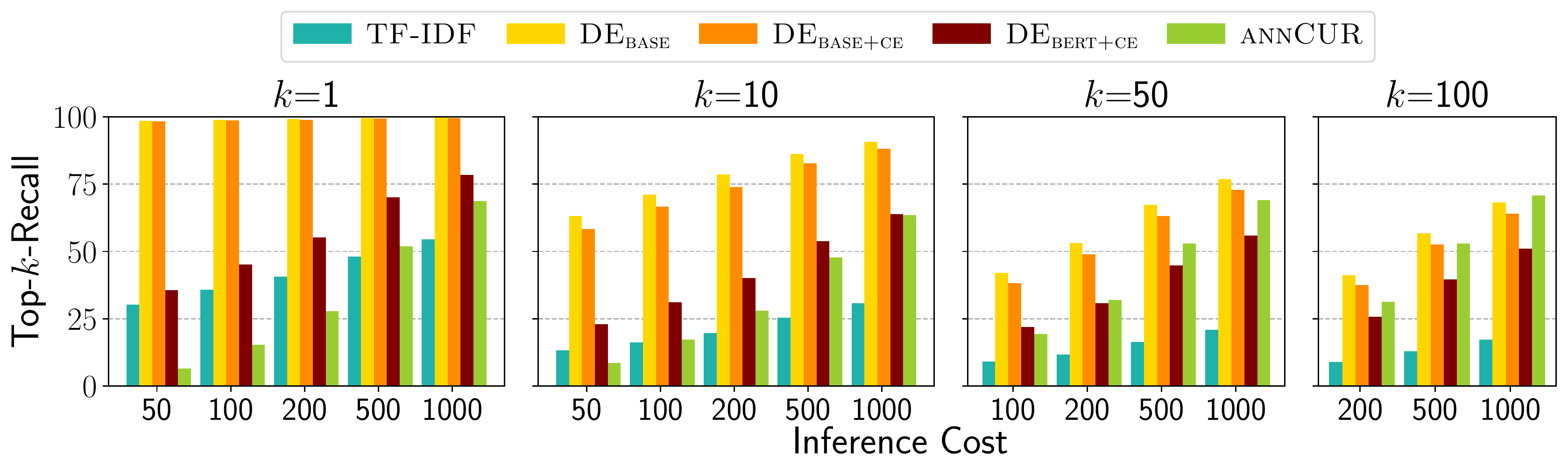}
    \caption{Top-$k$-Recall for \propMethod and baselines  when all methods operate under a fixed test-time cost budget. Recall 
    that cost is the number of CE calls made during inference
    for re-ranking retrieved items and, in case of \propMethod, it also includes CE calls to embed the test query by comparing with anchor items.
    }
    \label{fig:appendix_recall_at_same_cost_military_100}
\end{subfigure}
    \caption{Top-$k$-Recall results for domain=\military and $\queryTrainSize=100$}
\end{figure*}

\begin{figure*}[]
     \centering
     \begin{subfigure}[b]{0.95\textwidth}
        \centering
        \includegraphics[width=\textwidth]{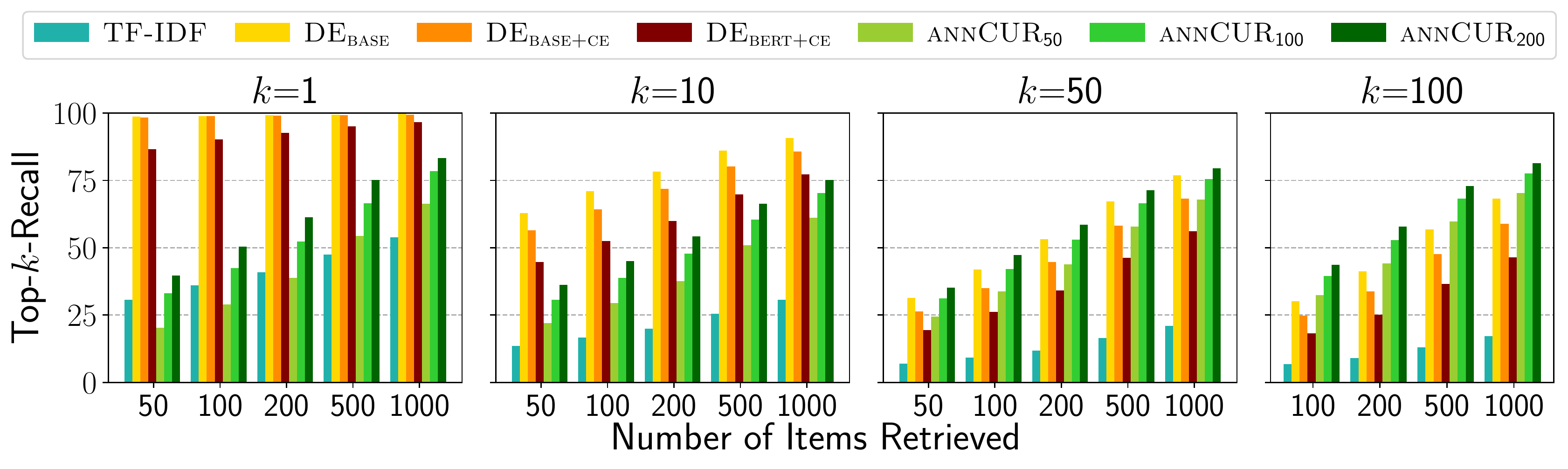}
        \caption{Top-$k$-Recall@$\nRetrievedItems$ for \propMethod and baselines when all methods retrieve and rerank 
        the same number of items $(\nRetrievedItems)$. The subscript $\nAnchorItems$ in $\propMethod_{\nAnchorItems}$ refers to
        the number of anchor items used for embedding the test query.}
        \label{fig:appendix_recall_at_same_k_retrieved_military_500}
    \end{subfigure}
     \hfill
     \begin{subfigure}[b]{0.95\textwidth}
    \centering
    \includegraphics[width=\textwidth]{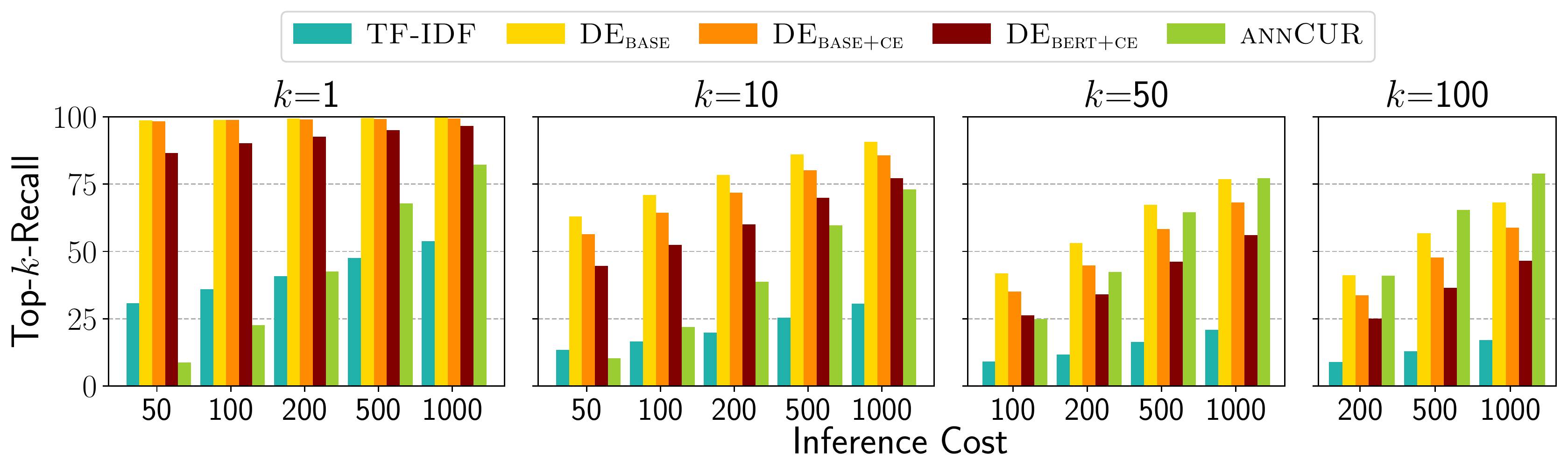}
    \caption{Top-$k$-Recall for \propMethod and baselines  when all methods operate under a fixed test-time cost budget. Recall 
    that cost is the number of CE calls made during inference
    for re-ranking retrieved items and, in case of \propMethod, it also includes CE calls to embed the test query by comparing with anchor items.
    }
    \label{fig:appendix_recall_at_same_cost_military_500}
\end{subfigure}
    \caption{Top-$k$-Recall results for domain=\military and $\queryTrainSize=500$}
\end{figure*}

\begin{figure*}[!ht]
     \centering
     \begin{subfigure}[b]{0.95\textwidth}
        \centering
        \includegraphics[width=\textwidth]{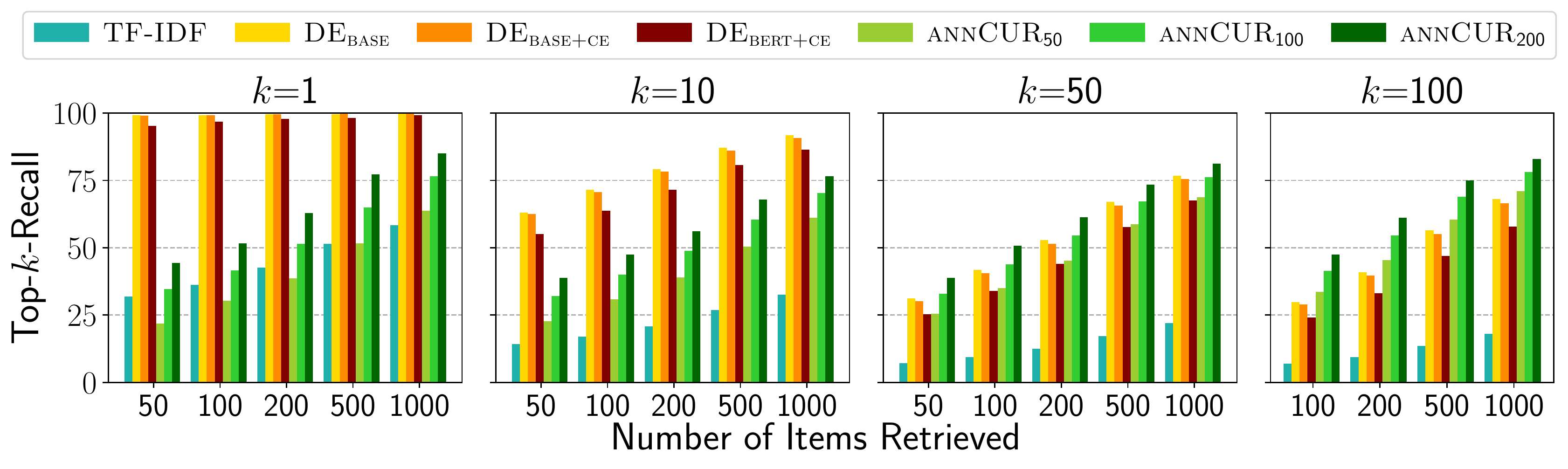}
        \caption{Top-$k$-Recall@$\nRetrievedItems$ for \propMethod and baselines when all methods retrieve and rerank 
        the same number of items $(\nRetrievedItems)$. The subscript $\nAnchorItems$ in $\propMethod_{\nAnchorItems}$ refers to
        the number of anchor items used for embedding the test query.}
        \label{fig:appendix_recall_at_same_k_retrieved_military_2000}
    \end{subfigure}
     \hfill
     \begin{subfigure}[b]{0.95\textwidth}
    \centering
    \includegraphics[width=\textwidth]{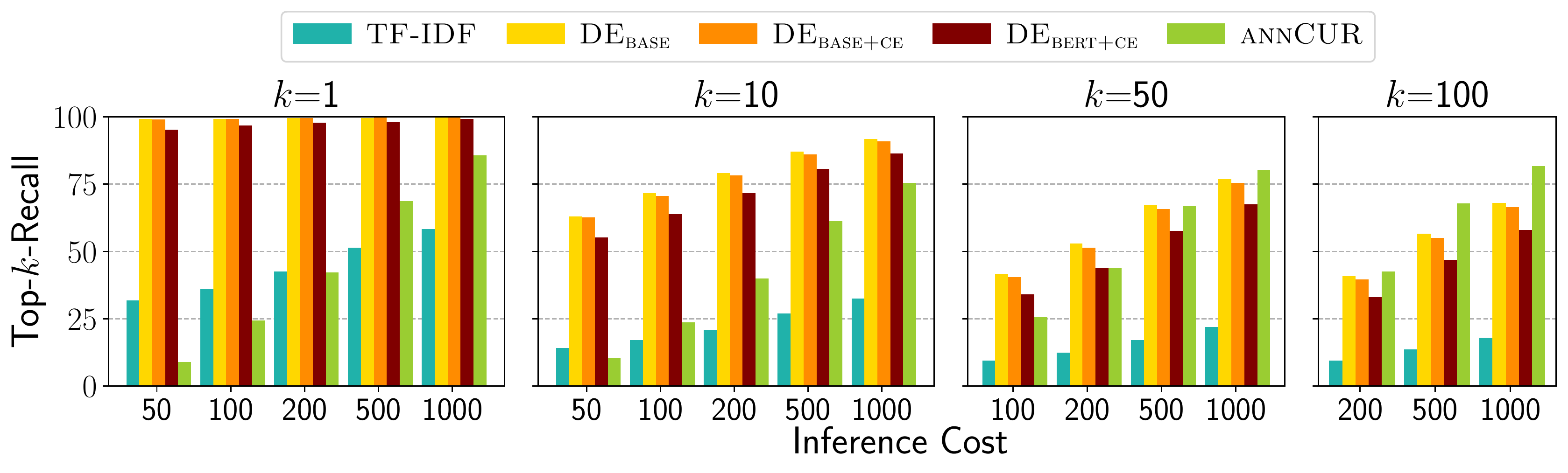}
    \caption{Top-$k$-Recall for \propMethod and baselines  when all methods operate under a fixed test-time cost budget. Recall 
    that cost is the number of CE calls made during inference
    for re-ranking retrieved items and, in case of \propMethod, it also includes CE calls to embed the test query by comparing with anchor items.
    }
    \label{fig:appendix_recall_at_same_cost_military_2000}
\end{subfigure}
    \caption{Top-$k$-Recall results for domain=\military and $\queryTrainSize=2000$}
    \label{fig:appendix_recall_military_2000}
\end{figure*}

For brevity, we show results for all top-$k$ values only for domain=\yugioh
in the main paper. For the sake of completeness and for interested readers,
we add results for combinations of top-$k$ values, domains, and training data size values.
Figure \ref{fig:appendix_recall_yugioh_100} - \ref{fig:appendix_recall_military_2000}
contain results for top-$k \in \{1, 10, 50, 100\}$, for domain \yugioh, \proWrestling, \doctorWho, \starTrek, \military,
and training data size $\queryTrainSize \in \{100, 500, 2000\}$. For \proWrestling, since the domain
contains 1392 queries, we use maximum value of $\queryTrainSize$ = 1000 instead of 2000.

\end{document}